%% file: templateArxiv.tex
\documentclass{article}

\usepackage{PRIMEarxiv}

\usepackage[utf8]{inputenc} 
\usepackage[T1]{fontenc}    
\usepackage{hyperref}       
\usepackage{url}            
\usepackage{booktabs}       
\usepackage{amsfonts}       
\usepackage{nicefrac}       
\usepackage{microtype}      
\usepackage{lipsum}
\usepackage{fancyhdr}       
\usepackage{graphicx}       

\graphicspath{{media/}}     

\usepackage{mathtools}
\usepackage{amsmath}
\usepackage{multirow}
\usepackage{xfrac}
\usepackage[caption=false]{subfig}
\usepackage{makecell}
\usepackage[table]{xcolor}
\usepackage{natbib}
\usepackage[title,page]{appendix}
\usepackage{amsthm}

\theoremstyle{definition}
\newtheorem{definition}{Definition}[section]

\DeclareMathOperator{\ACC}{ACC}
\DeclareMathOperator{\FPR}{FPR}
\DeclareMathOperator{\TPR}{TPR}

\DeclareMathOperator*{\argmin}{arg\,min}

\DeclareMathOperator{\outgrown}{OA}
\DeclareMathOperator{\EX}{\mathbb{E}}
\DeclareMathOperator{\Var}{Var}

\title{Quantifying With Only Positive Training Data
\thanks{\textit{\underline{Citation}}: 
\textbf{dos Reis, D. M., de Souto, M., de Sousa, E., \& Batista, G. (2020). Quantifying with only positive training data. arXiv preprint arXiv:2004.10356.}} 
}

\author{
    Denis M. dos Reis, Elaine P. M. de Sousa \\
    Instituto de Ci\^encias Matem\'aticas e de Computa\c{c}\~ao, Universidade de S\~ao Paulo\\
    Avenida Trabalhador S\~ao-carlense, 400, S\~ao Carlos - SP, Brazil\\
    \texttt{denismr@alumni.usp.br, parros@icmc.usp.br}
    \And
    Marc\'ilio C. P. de Souto \\
    LIFO Bat. 3IA, Universit\'e d'Orl\'eans\\
    Rue L\'eonard de Vinci – B.P. 6759, 45067 Orl\'eans Cedex 2, France\\
    \texttt{marcilio.desouto@univ-orleans.fr}
    \And
    Gustavo Batista \\
    Computer Science and Engineering, UNSW Sydney\\
    Sydney NSW 2052, Australia\\
    \texttt{g.batista@unsw.edu.au}

}

\begin{document}
\maketitle

\begin{abstract}
Quantification is the research field that studies methods for counting the number of data points that belong to each class in an unlabeled sample. Traditionally, researchers in this field assume the availability of labelled observations for all classes to induce a quantification model. However, we often face situations where the number of classes is large or even unknown, or we have reliable data for a single class. When inducing a multi-class quantifier is infeasible, we are often concerned with estimates for a specific class of interest. In this context, we have proposed a novel setting known as One-class Quantification (OCQ). In contrast, Positive and Unlabeled Learning (PUL), another branch of Machine Learning, has offered solutions to OCQ, despite quantification not being the focal point of PUL. This article closes the gap between PUL and OCQ and brings both areas together under a unified view. We compare our method, Passive Aggressive Threshold (PAT), against PUL methods and show that PAT generally is the fastest and most accurate algorithm. PAT induces quantification models that can be reused to quantify different samples of data. We additionally introduce Exhaustive TIcE (ExTIcE), an improved version of the PUL algorithm Tree Induction for $\mathfrak{c}$ Estimation (TIcE). We show that ExTIcE quantifies more accurately than PAT and the other assessed algorithms in scenarios where several negative observations are identical to the positive ones.
\end{abstract}


\input{sections/introduction}
\input{sections/background}
\input{sections/methods}
\input{sections/expsetup}
\input{sections/expeval}
\input{sections/discussion}
\input{sections/conclusion}
\input{sections/acknowledgements}
\input{sections/appendix}

\bibliographystyle{spbasic}
\bibliography{references}

\end{document}

%% file: sections/introduction.tex
\section{Introduction}
\label{intro}

Quantification is the task of estimating the prevalence (frequency) of the classes in an unlabeled sample of data, that is, counting how many data points belong to each class \citep{gonzalez2017review}. Several practical applications in diverse fields rely on quantifying unlabeled data points. In social sciences, quantification predicts election results by analyzing different data sources that support the candidates \citep{hopkins2010method}. In natural language processing, it assesses how probable is each meaning for a word \citep{Chan2006}. In entomology, it infers the local density of mosquitoes in a specific area covered by an insect sensor \citep{Chen2014}.

As is the case with classification, quantifiers generally learn from a labelled sample. In fact, one can achieve quantification by merely counting the output of a classification model. However, this approach produces suboptimal quantification performance \citep{tasche2016does}. The challenge of designing accurate counting methods has led to the establishment of quantification as a research area of its own, driven by a thriving community of researchers.

Although this community has been responsible for several novel quantification methods \citep{gonzalez2017review,maletzke2019dys}, they mainly focused on cases where there is plenty of labelled data for all classes. Furthermore, they assume that the set of classes is known a priori. However, depending on the problem at hand, we may be interested in counting observations that belong to a target class while not having substantial data from others.

For example, suppose we can use an intelligent sensor to count insects that belong to the \textit{Anopheles} mosquito genus, the vector of malaria, an infectious disease that affects more than 200 million people yearly. Even though we aim to count only a single class, the sensor will produce data points for other insect species in its vicinity. Considering that the number of insect species is estimated to be between six and ten million \citep{chapman2013insects}, it is unfeasible to build a dataset that reliably represents every non-target species. In another example, we may be interested in counting how many people in a social network would be interested in following a brand account. In this case, we only know which people already follow a certain brand, from which we can induce the behaviour of interested people. However, no data is available to model the behaviour of users not particularly interested in this brand.

In applications like these, we need a method capable of counting the target class while not directly modelling the behaviour of other classes. In other words, we cannot assume any available data to represent the behaviour of future observations that do not belong to the target class.

The previous problem is, in fact, a significant challenge that the Quantification community has overlooked. Indeed, to the best of our knowledge, we were the first to address this setting within the Quantification literature \citep{denisOCQ2018}. In our previous work, we introduced the task of One-class Quantification (OCQ). In OCQ, the goal is, from a training sample containing only data points belonging to the target class (positive data points), to induce a model that can estimate the prevalence of the positive class in a data sample containing unlabeled data points. 

As previously mentioned, we were the first researchers to define OCQ in the context of the Quantification community. However, in the broader context of Machine Learning, OCQ was not the first framework to tackle counting with only positive labelled data. Work in a research area called Positive and Unlabeled Learning (PUL) also developed methods that solve this problem with publications that go as far back as 2008 \citep{elkan2008}, under the task named Positive and Unlabeled Prior Estimation (PUPE). The main distinction between the methods proposed for PUPE and OCQ is that the former do not induce models that can be reapplied for several unlabeled test samples, while the latter does. Thus, to a great extent, both Quantification and PUPE share common goals. However, somewhat surprisingly, they have evolved as disparate academic fields.

Therefore, this paper contributes to building better awareness of how each area can enrich the other.
More specifically, in our previous work \citep{denisOCQ2018} we proposed Passive Aggressive Threshold (PAT) and One Distribution Inside (ODIn). We compared them solely against baseline and topline approaches under a narrow experimental setup (see Section~\ref{sec:exp:2}). In this paper, we extend our previous efforts by:

\begin{itemize}
    \item Reviewing some of the most relevant methods in PUPE literature in detail, thus unifying PUPE and OCQ literature;
    \item {Providing theoretical grounding and guaranties to PAT;}
    \item Extending our experimental setup to better understand the behaviour of the methods under varying circumstances (see Section~\ref{sec:exp:3});
    \item Comparing PAT against actual state-of-the-art approaches rather than baselines, according to quantification error and time required;
    \item Developing the Exhaustive TIcE (ExTIcE), a variation of the existing Tree Induction for $\mathfrak{c}$ Estimation \citep{bekker2018estimating}.
\end{itemize}

In our analysis, we discover that PAT outperforms all other methods tested in most settings we evaluated while being orders of magnitude faster.

However, by relying on scores as a proxy for data behaviour, PAT performance decreases when the target class overlaps to a great extent with other classes. To address this problem, we propose Exhaustive TIcE (ExTIcE), an extension of Tree Induction for $\mathfrak{c}$ Estimation (TIcE) \citep{bekker2018estimating}. ExTIcE can maintain quantification performance even with substantial overlap as long as, in some region of the feature space, there is little to no overlap. Although ExTIcE performs poorly in terms of time required for its computation, it serves the purpose of raising promising ideas for future work.

The following section provides a summary of essential concepts used throughout the remaining of our work. Section~\ref{sec:pupe} reviews the most prominent methods for PUPE and OCQ, including our proposals ExTIcE and PAT. We compare the reviewed methods according to the experimental evaluation described in Section~\ref{sec:exp:setup}, which led to the results presented and discussed in Section~\ref{sec:exp:eval}. Section~\ref{sec:discussion} discusses the strengths and limitations of the evaluated approaches and ways to compose them, opening some possibilities for future research. Finally, in Section~\ref{sec:conclusion} we conclude this article with a brief overview of our findings and prospects for future work.


%% file: sections/background.tex
\section{Background}
\label{sec:definitions}

This section introduces relevant definitions used throughout this work and clarifies the difference between classification and quantification tasks.

In Sections \ref{sec:def:class} and \ref{sec:def:scorer}, we explain classification and scoring, respectively, which are basic tools for several quantification techniques. Section~\ref{sec:def:quant} defines the quantification task, explains how it relates to classification, and demonstrates the limitation of achieving quantification through counting classifications. Section~\ref{sec:def:ocq} introduces One-class Quantification (OCQ), whose models do not rely on any expectations about the negative class and therefore forgo negative data. Section~\ref{sec:def:pupe} explains Positive And Unlabeled Prior Estimation (PUPE), which is similar to OCQ, albeit without requiring an explicitly reusable model. Finally,  Section~\ref{sec:def:diff_ocq_pupe} further differentiates OCQ from PUPE and how these differences impact the performance evaluation reported in the literature.

\subsection{Classification}
\label{sec:def:class}

In supervised learning, we are interested in learning from a \textit{training} sample $D = \{(\textbf{x}_1, y(\textbf{x}_1)) \allowbreak \dots, \allowbreak (\textbf{x}_n, y(\textbf{x}_n))\}$, where $\textbf{x}_i \in \mathcal{X}$ is a vector with $m$ attributes in the feature space $\mathcal{X}$, and $y: \mathcal{X} \mapsto \mathcal{Y}$ is a function that maps samples into a set of classes labels $\mathcal{Y}  = \{c_1, \ldots, c_l\}$. For the sake of readability, from now on we refer to $y(\textbf{x}_i)$ simply as $y_i$. Therefore, $D = \{(\textbf{x}_1, y_1), \allowbreak \dots, \allowbreak (\textbf{x}_n, y_n)\}$.

The classification objective is to correctly predict the class labels of observations in an unlabeled \textit{test} sample based on their feature values. A classifier is formalized in Definition~\ref{def:classifier}.

\begin{definition}
\label{def:classifier}
A \textbf{classifier} is a model $h$ induced from $D$ such that

\[h: \mathcal{X} \mapsto \mathcal{Y}\]

\noindent which aims to approximate the $y$ function.
\end{definition}

In classification, we usually assume that all observations are independent and identically distributed (\textit{i.i.d}) \citep{upton2014dictionary}. ``Identically distributed'' means that all observations, from either the training or test samples, share the same underlying distribution. ``Independently distributed'' means that the observations are independent of each other. In other words, the occurrence of one observation does not affect the probability of the occurrence of any other particular observation.

\subsection{Scorer}
\label{sec:def:scorer}

There are different mechanisms employed by classifiers to decide which class will be assigned to any given observation. We emphasize one that is frequently adopted for binary classification problems, that is, problems where $|\mathcal{Y}| = 2$. In binary classification, one of the two classes is denominated \emph{positive class} ($c_1 = \oplus$), while the other is denominated \emph{negative class} ($c_2 = \ominus)$. In this setting, one can induce a \textit{scorer} $S_h(\textbf{x})$, as formalized in Definition~\ref{def:scorer}

\begin{definition}
\label{def:scorer}
A \textbf{scorer} is a model $S_h$ induced from $D$ such that

\[S_h: \mathcal{X} \mapsto \mathbb{R} \]

\noindent which produces a numerical value called \textit{score} that correlates with the posterior probability of the positive class, that is $P(y=\oplus|\textbf{x})$.
Consequently, the greater the score, the higher the chance of $\textbf{x}$ belonging to the positive class.
\end{definition}

For classification purposes, if such a score is greater than a certain threshold $t_h$, the observation is classified as positive. Otherwise, it is classified as negative \citep{flach2012machine}. For the sake of brevity, we henceforth refer to scores of negative observations simply as \textit{negative scores}, and analogously refer to scores of positive observations as \textit{positive scores}. Such denominations are not to be confused with the sign of the numerical value of the scores. Given a scorer $S_h(\textbf{x})$, the classification task is fulfilled as follows:

\begin{equation*}
\label{eq:cl:by:scorer}
    h(\textbf{x}) =
    \begin{dcases}
    \oplus,     & \text{if } S_h(\textbf{x})> t_h\\
    \ominus,    & \text{otherwise}
    \end{dcases}
\end{equation*}

\subsection{Quantification}
\label{sec:def:quant}

Although quantification and classification share similar characteristics, the main one is the data representation, their objectives differ. A quantifier need not provide individual class predictions. Instead, it must assess the overall quantity of observations that belong to a specific class or a set of classes \citep{gonzalez2017review}. A quantifier is formally defined by Definition~\ref{def:quantifier}.

\begin{definition}
\label{def:quantifier}
A \textbf{quantifier} is a model induced from $D$ that predicts the prevalence of each class in a sample, such that

\[q: \mathbb{S}^{\mathcal{X}} \mapsto [0,1]^{l}\]

$\mathbb{S}^{\mathcal{X}}$ denotes the universe of possible samples from $\mathcal{X}$. For a given test sample $S \in \mathbb{S}^{\mathcal{X}}$, the quantifier outputs a vector $\hat{Q} = [\hat{p}_1, \allowbreak \dots \allowbreak, \hat{p}_l]$, where $\hat{p}_i$ estimates the prior probability for class $c_i$, such that $\sum_{j=1}^{l} \hat{p}_j = 1$. The objective is $[\hat{p}_1,\ldots,\hat{p}_l]$ to be as close as possible to the true prior ratios $[P(c_1),\ldots,P(c_l)]$ of the probability distribution from which $S$ was sampled.
\end{definition}

Similarly to classification, in quantification, we still assume that observations are sampled independently. Additionally, as the main task is to measure the prior probabilities of the classes in $S$, it is also assumed that the class distribution changes significantly from the training sample (which supports the induction of $q$) to the test sample $S$. Otherwise, a quantifier would not be needed.

One straightforward way of achieving quantification is to count the predictions produced by a classifier. This method is called Classify and Count (CC) \citep{Forman2005}. Naturally, performing CC with a perfect classifier always produces a perfect quantification. However, accurate quantifiers do not necessarily need to rely on accurate classifiers. Since our objective is purely to count how many observations belong to each class, misclassifications can nullify each other, as illustrated in Table~\ref{tab:cc:example}.

\begin{table}[htb]
\centering
\begin{tabular}{ll|ccc}
\hline

\hline
\multicolumn{2}{c}{}&\multicolumn{2}{c}{Prediction}& \\
\multicolumn{2}{c}{}&Positive&Negative&Total\\
\hline
\multirow{2}{*}{Actual}& Positive & $830$ & $170$ &\multicolumn{1}{|c}{$1000$}\\
& Negative & $170$ & $330$ & \multicolumn{1}{|c}{$500$}\\
\hline
\multicolumn{1}{c}{} & \multicolumn{1}{c}{Total} & \multicolumn{1}{c}{$1000$} & \multicolumn{    1}{c}{$500$} & \multicolumn{1}{c}{$1500$}\\
\hline

\hline
\end{tabular}
\caption{Confusion matrix of a toy classification model that achieves 77\% accuracy given a test sample. Although the model is not a perfect classifier, it provides perfect quantification in the presented scenario: it predicts the number of positive observations to be 1,000, which is the correct amount -- false positives and false negatives cancel out.}
\label{tab:cc:example}
\end{table}

\begin{figure}[htb]
    \centering
    \includegraphics[width=.8\columnwidth]{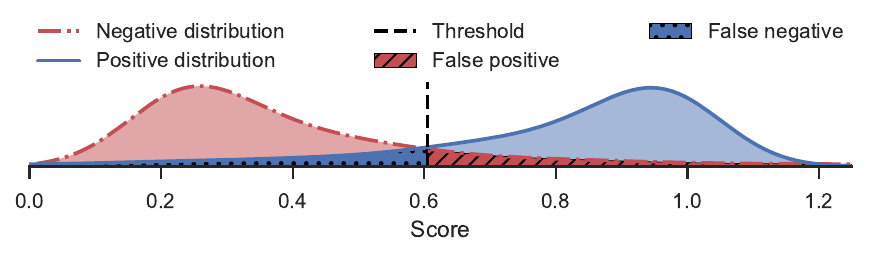}
    \caption{A quantifier that relies on a scorer-based classifier. We show the probability density functions of the scores for both classes. The density functions are scaled to depict the proportion of the classes. For the given threshold, we have a nearly perfect quantification, albeit classification is not perfect. The classification errors nullify each other~\citep{denisOCQ2018}.}
    \label{fig:bincut}
\end{figure}

As in Table~\ref{tab:cc:example}, Figure~\ref{fig:bincut} illustrates a scenario where we obtain perfect quantification regardless of imperfect classification, given a classification model based on a scorer. This illustration will come in handy to visually understand the systematic error of CC, afterwards.

Despite CC providing perfect quantification in specific test conditions, it is systemically flawed for any classifier that does not consistently achieve perfect classification. Perfect classifiers are rarely achievable for real-world applications. For illustration purposes, consider a case of binary quantification. Let $\hat{p}$ be the estimated proportion of the positive class in an unlabeled test sample, while $p$ is the true positive class ratio in the unlabeled test sample. In CC, $\hat{p}$ is estimated as follows:

\[
    \hat{p} = \frac{\textrm{unlabeled instances classified as positive}}{\textrm{unlabeled instances}}
\]

Observe that we can decompose $\hat{p}$ in terms of how many positive observations were correctly classified and how many negative observations were incorrectly classified, even though these values are not obtainable without access to true labels.

\[
    \hat{p} = \frac{\textrm{true positives } + \textrm{false  positives}}{\textrm{unlabeled instances}}
\]

To put $\hat{p}$ in a probabilistic perspective, let $P_h(\hat{\oplus}|\ominus)$ be an alias for $P(h(\textbf{x}) = \oplus | y(\textbf{x}) = \ominus)$, which the classifier's False Positive Rate (FPR). In other words, it is the proportion of negative observations that are wrongly classified as positive. Analogously, let $P_h(\hat{\oplus}|\oplus) \allowbreak$ be an alias for $P(h(\textbf{x}) = \ominus | y(\textbf{x}) = \oplus)$, which is the classifier's True Positive Rate (TPR). In other words, it is the proportion of positive observations that are correctly classified as such. In this context, $\hat{p}$ can be defined as:

\begin{equation}
\label{eq:CC:desc}
    \hat{p} = P(h(\textrm{x})=\oplus) = pP_h(\hat{\oplus}|\oplus) + (1 - p)P_h(\hat{\oplus}|\ominus)
\end{equation}

From the previous equation, we can derive that the absolute quantification error $|\epsilon_{\textrm{CC}}|$ caused by CC is:

\begin{align}
\label{eq:abserr}
\begin{split}
    \left|\epsilon_{\textrm{CC}}\right|
    &= \left|\hat{p} - p\right| \\
    &= \left|pP_h(\hat{\oplus}|\oplus) + (1 - p)P_h(\hat{\oplus}|\ominus) - pP_h(\hat{\oplus}|\oplus) - pP_h(\hat{\ominus}|\oplus)\right|\\
    &= \left|(1 - p)P_h(\hat{\oplus}|\ominus) - pP_h(\hat{\ominus}|\oplus)\right|\\
    &= \left|pP_h(\hat{\ominus}|\oplus) - (1 - p)P_h(\hat{\oplus}|\ominus)\right|
\end{split}
\end{align}

\noindent where $P_h(\hat{\ominus}|\oplus) \allowbreak$ is an alias for $P(h(\textbf{x}) = \ominus | y(\textbf{x}) = \oplus)$, which is the classifier's False Negative Rate (FNR) or, in other words, the proportion of positive observations that are wrongly classified as negative.

From Equation~\ref{eq:abserr}, observe that the error relates to the absolute difference between the hatched areas (false positive and false negative) in Figure~\ref{fig:bincut}. Intuitively, this means that, for a score-based quantifier, it is enough to select a threshold that causes the number of false-negative observations to be the same as the number of false-positive observations. However, those values depend on the true-positive ratio $p$, which is the variable we want to estimate in the first place, thus making this method of choosing a threshold impracticable.
Observe that if we do not apply the absolute function, we can easily express $\epsilon_{CC}$ as a linear function of $p$:

\begin{align}
\label{eq:cc:error}
\begin{split}
    \epsilon_{\textrm{CC}}(p) &= pP_h(\hat{\ominus}|\oplus) - (1 - p)P_h(\hat{\oplus}|\ominus)\\
        &= \left(P_h(\hat{\ominus}|\oplus) + P_h(\hat{\oplus}|\ominus)\right)p - P_h(\hat{\oplus}|\ominus) \\
        &= \alpha p + \beta
\end{split}
\end{align}

\begin{figure}[htb]
    \centering
    \includegraphics[width=.5\columnwidth]{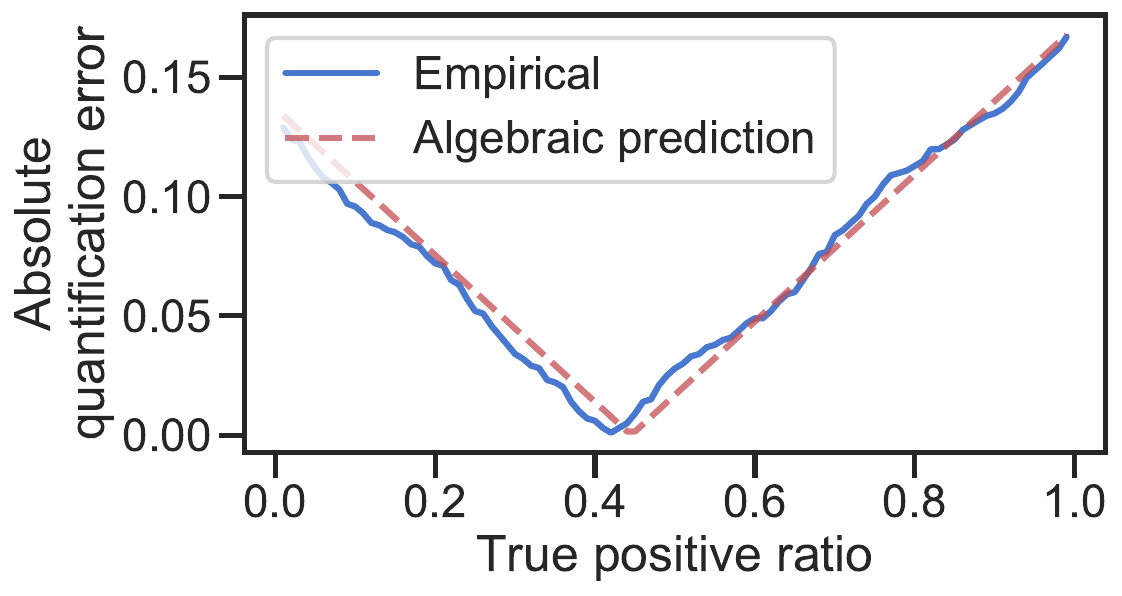}
    \caption{Experimental performance of a quantifier that relies on a scorer-based classifier for several true positive ratios. The continuous blue curve is the empirical absolute quantification error (AQE) produced by the model. The red dotted curve is the algebraic prediction for the absolute AQE (Equation~\ref{eq:cc:error}) given estimates of FPR and TPR obtained via 10-fold cross-validation. The classifier is a Random Forest with 100 trees. The score is given by how many votes the positive class received. Dataset (Insects v2, described in Section~\ref{sec:datasets}) includes flight information for female \textit{Aedes aegypti} (positive class) and female \textit{Culex quinquefasciatus} (negative class).}
    \label{fig:cc:demo}
\end{figure}

This implies that $|\epsilon(p)_{\textrm{CC}}|$, the absolute quantification error generated by CC, grows linearly when the actual positive class ratio $p$ is under or above a certain value for which quantification should be perfect. This effect is true for any classifier whose either $P_h(\hat{\oplus}|\ominus)$ or $P_h(\hat{\ominus}|\oplus)$ is not null. Figure~\ref{fig:cc:demo} illustrates such an effect with a real dataset.

\begin{figure}[htb]
    \centering
    \includegraphics[width=.8\columnwidth]{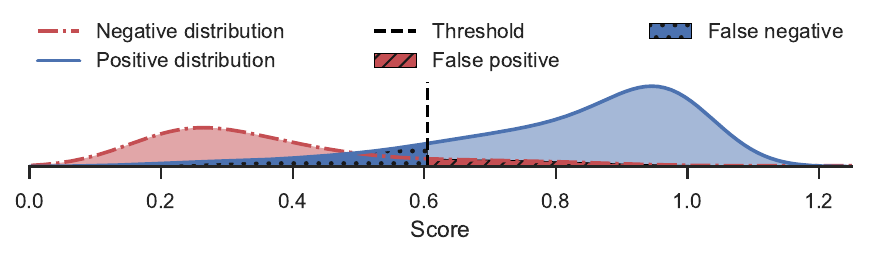}
    \caption{Illustration of a quantifier that relies on a scorer-based classifier. This illustration shows the probability density functions of the scores for both classes. The density functions are scaled to depict the proportion of the classes. Contrary to Figure~\ref{fig:bincut}, the quantification is not perfect with the current proportion of positive observations, even though the classification threshold and the probability distributions for each class, taken individually, are all the same~\citep{denisOCQ2018}.}
    \label{fig:bincut:err}
\end{figure}

Figure~\ref{fig:bincut:err} further illustrates a change of $p$ on the density function of scores in a score-based classifier. Compared to Figure~\ref{fig:bincut}, we can notice that the area related to false-positive observations shrunk down, while the area related to false-negative observations expanded as $p$ gets bigger. In general, if the proportion of positive observations is greater than the one that causes perfect quantification, the predicted positive ratio is underestimated since the number of false negatives becomes greater than the number of false positives. Likewise, if the proportion of positive observations is lower than the one that causes perfect quantification, the predicted positive ratio is overestimated. We point the interested reader to \cite{tasche2016does} for a thorough investigation of quantification limitations without adjustments.

A similar systematic error pattern is found if we extend our analysis on binary CC to the multiclass scenario.

Although most quantification algorithms rely, at some point, on classifiers or scorers, there are several ways to minimize the systematic error. In the binary case, if we rewrite Equation~\ref{eq:CC:desc} to isolate the true positive ratio $p$, we have:

\begin{equation}
    \label{eq:acc:source}
    p = \frac{\hat{p} - P(\hat{\oplus}|\ominus)}{P(\hat{\oplus}|\oplus) - P(\hat{\oplus}|\ominus)}
\end{equation}

With Equation~\ref{eq:acc:source} we conclude that if we know the actual values of TPR and FPR, we can calculate $p$ as a function of $\hat{p}$. That is, we can derive the actual proportion of positive observations $p$ from the biased $\hat{p}$ estimated by Classify and Count. This is the principle of Adjusted Classify and Count (ACC) \citep{Forman2005}, which is defined in the following equation, where $\widehat{\TPR}$ and $\widehat{\FPR}$ are estimates of $\TPR$ and $\FPR$, respectively:

\begin{equation*}
\ACC(\hat{p}, \widehat{\TPR}, \widehat{\FPR}) = \min\left\{1, \frac{\hat{p} - \widehat{\FPR}}{\widehat{\TPR} - \widehat{\FPR}}\right\}
\end{equation*}

As ACC comes from Equation~\ref{eq:acc:source}, it produces perfect quantification when the estimates of FPR and TPR are both correct. However, $\widehat{\FPR}$ and $\widehat{\TPR}$ are typically empirically estimated with labelled training data and procedures such as $k$-fold cross-validation. Therefore, such estimates may be imperfect, mainly when validation sets are limited or the class distribution is imbalanced.

\subsection{One-class Quantification}
\label{sec:def:ocq}

Now, let us turn our attention to the context in which we want to quantify only one target class (positive class). The negative class is a mixture of distributions that comprises everything that does not belong to this target class. Each component in this mixture is a negative sub-class. The data we have available for negative sub-classes constitute our partial knowledge about the negative class.

One problem with typical quantification is that if there is an exceedingly large number of negative sub-classes, the ones for which we have data might not be enough to model the general behaviour of the negative class reliably.

There are cases where it is difficult to guarantee that an observation belongs to the negative class. For example, suppose that we sell a product online and track our customers' preferences via their social media profiles. Our customers can be used as positive training data to identify who might be interested in purchasing our product. On the other hand, gathering data for the negative class is not as trivial. If we randomly sample online social media profiles, the resulting set would contain people not interested in the product and potential customers. An explicit data gathering for the negative class could involve an online poll, which is time-consuming and can still generate potentially biased data.

In a second example, suppose that we want to count the number of infected people with a disease in a population. Due to procedure costs, people may test for a disease only if they are suspected of having it. In that case, while we can have a sample of positively tested people for such a disease, our data for people who were negatively tested may be severely lacking and biased. In such a case, a random sample of people would include people who are not infected and people who are infected but were never diagnosed.

Suppose we are interested in quantifying only the positive class, and we cannot have a reliable representation of the negative class. In that case, we may need to rely solely on positive training data to induce a quantification model. 

One-class Quantification (OCQ) is the task of inducing a quantification model with only positive data, as formalized in Definition~\ref{def:quantifierOC} \citep{denisOCQ2018}.

\begin{definition}
\label{def:quantifierOC}
A \textbf{one-class quantifier} is a quantification model induced from a single-class dataset, in which all available labeled examples belong to the same class, say the positive one, $D^{\oplus} = \{(\textbf{x}_1, \oplus), \dots, \allowbreak (\textbf{x}_n, \oplus)\}$, and

\[q^{\oplus}: \mathbb{S}^{\mathcal{X}} \longrightarrow [0,1]\]

The one-class quantifier outputs a single probability estimate $\hat{p} \in [0,1]$ of the positive class prevalence. Notice, however, that $q^{\oplus}$ operates over $\mathbb{S}^{\mathcal{X}}$, \textit{i.e.}, a sample with all occurring classes. 
\end{definition}

Excluding the explicit objective of inducing a model and disregarding training data afterwards, OCQ shares the same purposes of Positive and Unlabeled Prior Estimation, detailed in the next section.

\subsection{Positive and Unlabeled Prior Estimation}
\label{sec:def:pupe}

Positive and Unlabeled Prior Estimation (PUPE) is a task derived from Positive and Unlabeled Learning (PUL). The main task of the latter is akin to classification. To better explain PUPE, we first briefly introduce PUL.

\begin{figure}[htb]
    \centering
    \includegraphics[width=.4\columnwidth]{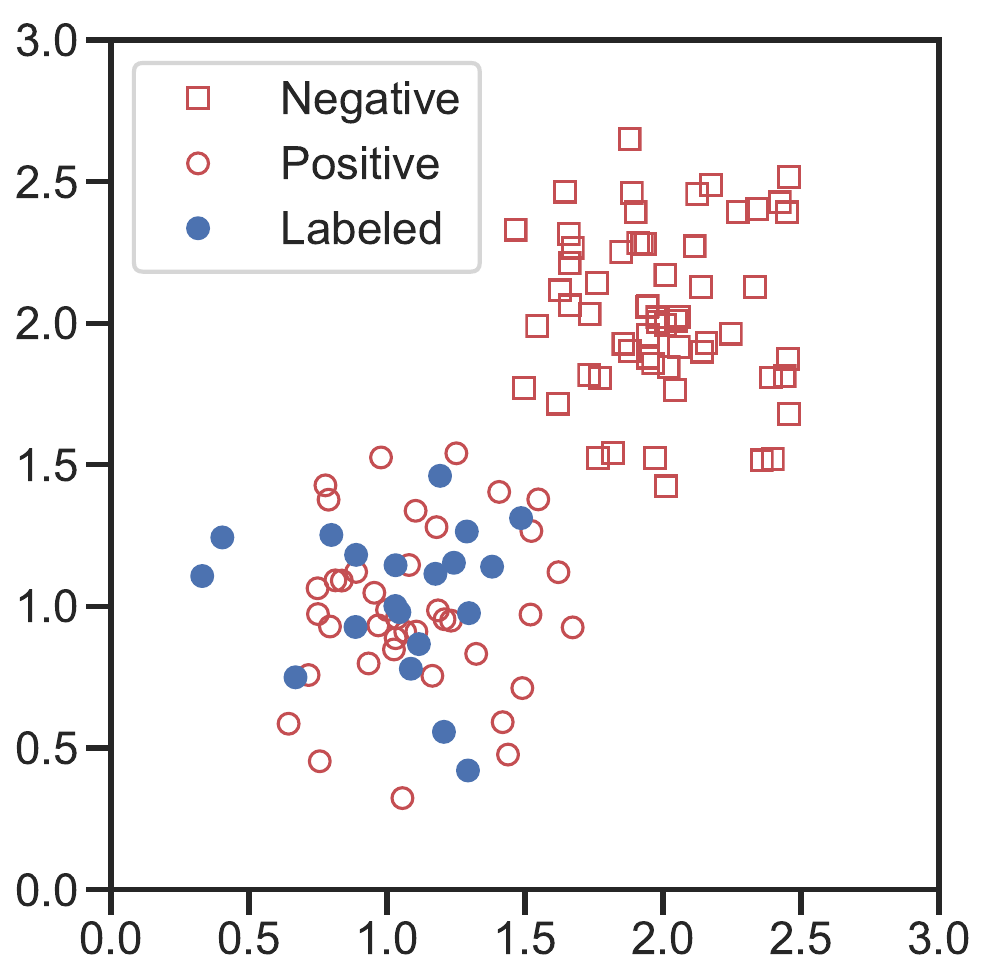}
    \caption{The general setting of Positive and Unlabeled Learning in a two-dimensional feature space. The filled (blue) circles correspond to the exclusively positive data sample, and the unfilled shapes correspond to the unlabeled data sample. In the unlabeled data sample, circles are positive observations, and squares are negative observations. However, such labels in the unlabeled sample are not provided to the PU task at hand.}
    \label{fig:elkan:before}
\end{figure}

In the general case of PUL \citep{elkan2008}, we are provided with two data samples. One of such samples, $L$, contains only positive (and therefore \textbf{l}abeled) observations, whereas the other, $U$, contains \textbf{u}nlabeled observations that can be either positive or negative. The objective is to infer the individual labels of the observations in the unlabeled sample. Figure~\ref{fig:elkan:before} illustrates the general setting of PUL.

Observe that the basic description of PUL does not pose explicit restrictions regarding the proportion of the classes in the unlabeled data. However, possessing such a statistic makes the labelling task an easier problem \citep{elkan2008}. If the labelling is based on a scorer, we can use the number of positive observations to set a classification threshold. Unfortunately, the number of positive observations in an unlabeled sample is not readily available, although it can be estimated. In that sense, Positive and Unlabeled Prior Estimation (PUPE) is a sub-task that has the sole objective of predicting the proportion of the classes, which could eventually support labelling.

A common assumption across different pieces of work on PUL and PUPE is that the labeled sample is ``selected completely at random'' from the pool of positive examples. More specifically, such an assumption states that each positive observation has a constant probability of $\mathfrak{c}$ of being labeled \citep{elkan2008}. Consider $s$ a function that annotates whether a positive observation is labeled, as follows:

\begin{equation*}
    s(\textbf{x}) = \begin{cases}
        1 & \textrm{if } y(\textbf{x}) = \oplus \textrm{ and \textbf{x} is labeled} \\
        0 & \textrm{otherwise}
    \end{cases}
\end{equation*}

In such a case, the assumption specifies that

\begin{align}
\label{eq:puq:c}
\begin{split}
    \mathfrak{c} &= P(s(\textbf{x}) = 1 | \textbf{x}, y(\textbf{x}) = \oplus)\\
    &= P(s(\textbf{x}) = 1 | y(\textbf{x}) = \oplus)
\end{split}
\end{align}

\noindent that is, the probability of $s(\textbf{x}) = 1$ is a constant $\mathfrak{c}$ for any $\textbf{x}$ belonging to the positive class. Note that, by definition, $P(s(\textbf{x}) = 1|y(\textbf{x}) = \ominus) = 0$. By applying the Bayes Theorem, also follows that

\begin{align*}
    \mathfrak{c} &=
    P(s(\textbf{x}) = 1 | y(\textbf{x}) = \oplus)\\
    &= \frac{P(s(\textbf{x}) = 1, y(\textbf{x}) = \oplus    )}{P(y(\textbf{x}) = \oplus)}
\end{align*}

\noindent and

\begin{align*}
    1 &= 
    P(y(\textbf{x}) = \oplus | s(\textbf{x}) = 1)\\
    &=
    \frac{P(s(\textbf{x}) = 1, y(\textbf{x}) = \oplus)}{P(s(\textbf{x}) = 1)}
\end{align*}

\noindent from which follows \citep{elkan2008}

\begin{equation}
\label{eq:puq:c2}
    P\left(s(\textbf{x}) = 1\right) = \mathfrak{c}P\left(y(\textbf{x}) = \oplus\right)
\end{equation}

In a simplification, the labelled sample is a uniform sample from all available positive observations. More importantly, this assumption and how the algorithms exploit it underlines that the labelled sample and the positive observations from the unlabeled sample share the same probability distribution. Therefore,

\begin{align*}
    P(\textbf{x}|s(\textbf{x}) = 1) &= P(\textbf{x}|s(\textbf{x}) = 0, y(\textbf{x}) = \oplus) \\ &= P(\textbf{x}|y(\textbf{x}) = \oplus)
\end{align*}

We note that this assumption is also made in OCQ methods since they aim to induce a model that estimates the probability distribution of the positive class. Despite this similarity of assumption, there are differences between OCQ and PUPE that are better described in the next section.

\subsection{Differences between OCQ and PUPE}
\label{sec:def:diff_ocq_pupe}

Having described OCQ and PUPE, we stress that, from a practical perspective, algorithms from both research areas are capable of solving the same set of problems interchangeably. Therefore, direct comparisons between the methods of such areas are due. However, while both methods \textit{can} solve the same set of problems, there is an essential distinction between the problems they \textit{aim} to solve. 
PUPE describes the task as containing exactly two samples: there is no particular interest in modelling a single model that can quantify several test samples. Such a description influenced the development of PUPE techniques. As a result, all of the examined techniques rely on transductive learning at all stages of the quantification process: they do not produce a single model that can be reused. Therefore, we need to perform a costly process for each test samples that we need to evaluate.

On the other hand, OCQ methods create a model that estimates the distribution of the positive class and quantifies any given sample at a later time. As we show in this article, this perspective to the problem provided OCQ techniques with a sizable advantage in terms of time needed to process a large number of test samples over PUPE techniques.

We also note that in the literature on PUPE, the task is often to estimate either $\mathfrak{c}$ or $P(y(\textbf{x}) = \oplus)$, whereas in OCQ we are interested in estimating $p$. Note $P(y(\textbf{x}) = \oplus)$ is the chance of one observation belonging to the positive class considering \textit{both} labeled data and unlabeled data. Also, recall that $\mathfrak{c} = P(s(\textbf{x}) = 1|y(\textbf{x}) = \oplus)$, that is, $\mathfrak{c}$ is the ratio of labeled data to unlabeled positive data. Both probabilities listed depend on the labeled set, which is intended for training.

Meanwhile, in a conversion to the PUPE terminology, $p = P(y(\textbf{x}) = \oplus | s(\textbf{x}) = 0)$, that is, the proportion of the positive observations considering \textit{only the unlabeled set}. This divergence is reflected in how the literature presents its experimental results. We highlight that by measuring the error of estimates of either $\mathfrak{c}$ or $P(y(\textbf{x}) = \oplus)$, the value obtained is highly influenced by the number of labelled observations (which are training data). On the other hand, the training data size does not influence evaluation measurements based on $\hat{p}$. Thus, given our discussion, we argue that one should adopt evaluation metrics based on $\hat{p}$ to measure the performance of methods in either OCQ or PUPE.

We can convert the estimation $\hat{\mathfrak{c}}$ to $\hat{p}$ according to the following equation:

\begin{equation*}
    \hat{p} = \min\left\{1, 
    \frac{\hat{\mathfrak{c}}^{-1}|L| - |L|}{|U|}
    \right\}
\end{equation*}

\noindent where $|L|$ is the number of labelled observations and $|U|$ is the number of unlabeled observations. The $\min$ function in the expression limits the result since the predicted $\hat{\mathfrak{c}}$ can have a corresponding $\hat{p}$ over one, which would be meaningless for quantification. Observe that $\hat{p}$ is inversely proportional to $\hat{\mathfrak{c}}$.

Finally, we emphasize that although the general assumption of both OCQ and PUPE is that the negative class cannot be estimated from labelled data, the distribution of the negative class does not fully contain the distribution of the positive class. In other words, the positive class is at least partially separable from the negative class. Algorithms may impose stricter versions of this assumption to achieve quantification successfully. For instance, Elkan's algorithm requires a clear separation between negative observations and positive observations.




%% file: sections/methods.tex
\section{Methods for Quantifying with Only Positive Labeled Data}
\label{sec:pupe}

In this section, we review and describe the more prominent methods in PUPE literature, highlighting key aspects of their rationale and implications in practical use, in addition to a new method, ExTIcE, that we propose in this paper. We do our best to simplify the rationale behind each method and offer a more intuitive and approachable explanation that unveils the uniqueness of each algorithm.

{
Finally, we revisit our previously introduced OCQ method, PAT, providing before unexplored theoretical guarantees and new insights regarding parameter tuning.
}

\subsection{Elkan (EN)}

To the best of our knowledge, \citet{elkan2008} were the first to explicitly tackle the prior estimation problem in Positive and Unlabeled Learning as a separate task. They introduce three techniques to estimate $\mathfrak{c}$, one of which, henceforth called Elkan's method (EN), is their recommended choice. The rationale of this method directly derives from Equation~\ref{eq:puq:c}. Precisely, the technique tries to estimate $\mathfrak{c} = P(s(\textbf{x}) = 1 | y(\textbf{x}) = \oplus)$ with the two following steps:

In the first step, using both unlabeled $U$ and labeled $L$ datasets together, we train a classification model capable of producing \textit{calibrated} scores (probabilities), where the class feature is whether the observation belongs to $L$ or not. In other words, the classifier aims to predict $s(\textbf{x})$ rather than $y(\textbf{x})$. As the model is a calibrated scorer, it estimates $P(s(\textbf{x}) = 1|\textbf{x})$.

In the second step, in order to estimate $P(s(\textbf{x}) = 1 | y(\textbf{x}) = \oplus)$ and therefore $\mathfrak{c}$, EN uses $L$ as a proxy for the condition $y(\textbf{x}) = \oplus$ of the aforementioned probability. It  averages all probabilities obtained for the observations in $L$ as follows:

\begin{equation*}
    \hat{\mathfrak{c}} = |L|^{-1}\sum_{\textbf{x}_i \in L}\hat{P}(s(\textbf{x}_i) = 1|\textbf{x}_i)
\end{equation*}

\begin{figure}[htb]
    \centering
    \includegraphics[width=.4\columnwidth]{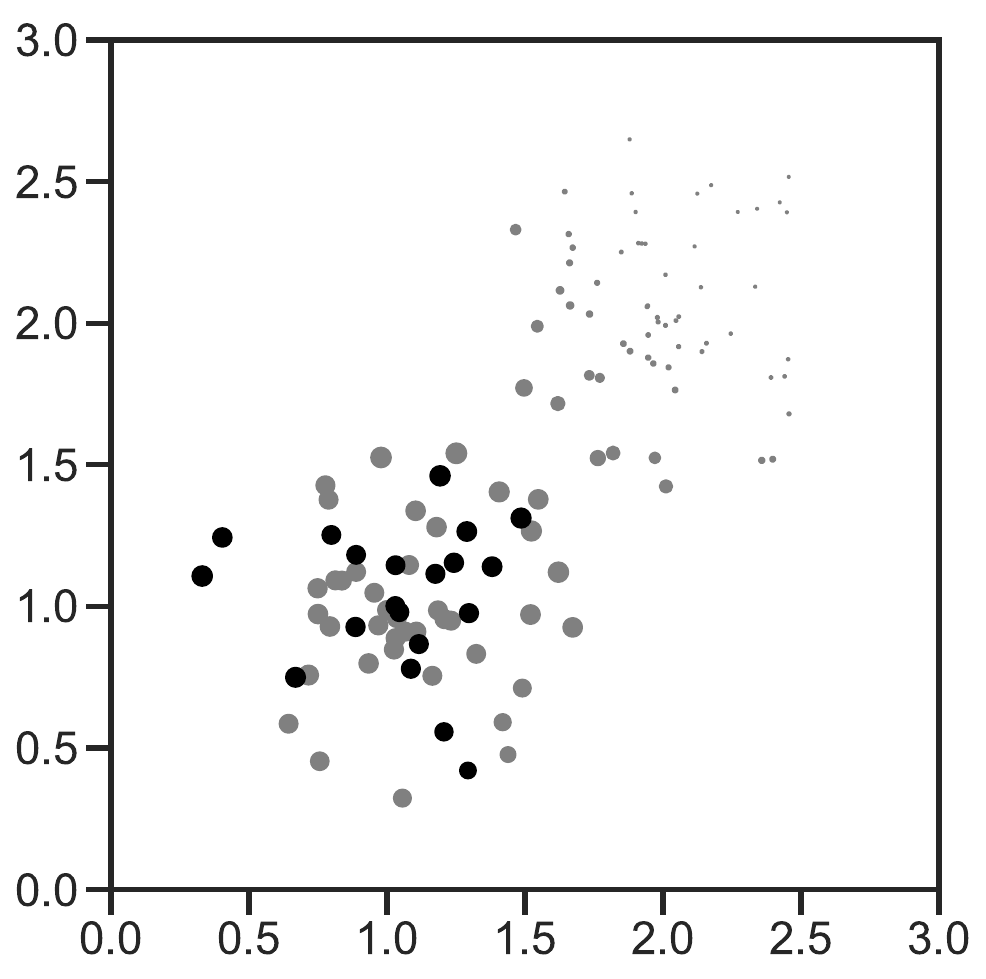}
    \caption{Illustration of Elkan's algorithm. The points on the plot correspond to a uniform sample of $10\%$ of the data used to run the algorithm. The size of the points correlates linearly with the estimated probability of the observation being labeled. Gray dots correspond to unlabeled data (unused to estimate $\mathfrak{c}$ once the model is trained). Black dots are labeled positive observations. $\hat{\mathfrak{c}} = 30\%$ is the average of the estimated probabilities of black dots being labeled. In contrast, actual $\mathfrak{c}$ is $33\%$.}
    \label{fig:elkan:after}
\end{figure}

Figure~\ref{fig:elkan:after} exemplifies Elkan's algorithm on the same dataset that generated Figure~\ref{fig:elkan:before}. We make two observations based on these figures. First, positive observations, either labeled or unlabeled, share similar sizes in Figure~\ref{fig:elkan:after}. Indeed, as they have the same probability distribution, they also share the same area in the feature space uniformly. In such a case, where features are useless to distinguish labeled observations from positive but unlabeled ones, the best possible estimation for the probability of any single positive observation being labeled is the proportion of labeled observations in the \textit{shared} space, therefore $\mathfrak{c}$ (see Equation~\ref{eq:puq:c}).

The second important aspect to note, in Figure~\ref{fig:elkan:after}, is that as a negative observation gets farther the positive cluster, it also gets smaller. This happens because they get farther from \textit{labeled} observations, which are the classification target  for the model induced. This remark raises the question of what would happen if there were further spatial overlap between the classes. Notice that EN estimates $\mathfrak{c}$ by averaging $\hat{P}(s(\textbf{x}_i) = 1|\textbf{x}_i)$ for all $x_i \in L$. This works on the assumption that

\begin{align*}
\hat{P}(s(\textbf{x}_i) = 1|\textbf{x}_i) 
&= \hat{P}(s(\textbf{x}_i) = 1|\textbf{x}_i, y(\textbf{x}_i) = \oplus) \hspace{0.5cm}\\
&= \hat{P}(s(\textbf{x}_i) = 1| y(\textbf{x}_i) = \oplus) \hspace{0.5cm}\\
&\forall \hspace{0.1cm} \textbf{x}_i \in L
\end{align*}

While it is true that $y(\textbf{x}_i) = \oplus$ for every observation $\textbf{x}_i$ in $L$, we emphasize that the classification model learns how to estimate $\hat{P}(s(\textbf{x}_i) = 1|\textbf{x}_i)$, \textit{not} $\hat{P}(s(\textbf{x}_i) = 1|\textbf{x}_i, y(\textbf{x}_i) = \oplus)$. The true value of the former probability is given according to the following equation:

\begin{equation*}
    P(s(\textbf{x}_i) = 1|\textbf{x}_i) = P(y(\textbf{x}_i) = \oplus)
    P(s(\textbf{x}_i) = 1|y(\textbf{x}_i) = \oplus)
\end{equation*}

By providing the classifier only with instances from $L$, EN implicitly assumes that $P(y(\textbf{x}_i) = \oplus) = 1$, whereas it may not be the case. Indeed, $P(s(\textbf{x}_i) = 1)$ will be significantly lower than $P(s(\textbf{x}_i) = 1|y(\textbf{x}_i) = \oplus)$ when there is overlap between the classes, since in such cases $P(y(\textbf{x}_i) = \oplus) \ll 1$. For this reason, when there is overlap between the classes, EN  underestimates $\hat{\mathfrak{c}}$ and therefore overestimates $\hat{p}$. As we show in the next sections, newer algorithms handle the possibility of class overlap better than EN by different means.

\subsection{PE, pen-L1 and ODIn}

{
A number of methods estimate $\hat{p}$ by partially matching the distribution of the positive class with the distribution of the whole data. The distribution of the positive class is estimated from $L$ while the distribution of the whole data is estimated from $U$. The difference between methods is on how the distributions are estimated, the matching criteria, and which method is used to choose a value for $\hat{p}$ that satisfies the such a criteria. In general, the whole process is illustrated in Figure~\ref{fig:partial:matching}.

\begin{figure}[htb]
    \centering
    \includegraphics[width=.8\columnwidth]{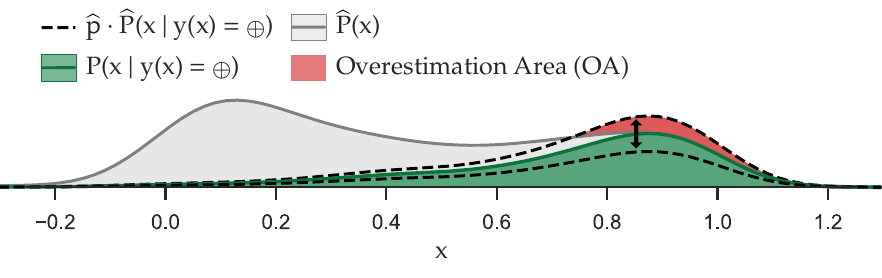}
    \caption{Illustration of partial matching. The dotted curves represent the effect of varying $\hat{p}$ in $\hat{p}P(\textbf{x}|y(\textbf{x})=\oplus)$. Ideally, $\hat{p}$ is chosen so that the dotted curve coincides with the curve of $P(\textbf{x} | y(\textbf{x}) = \oplus)$, which is the actual distribution of the positive class, and therefore the Overestimation Area (OA) is zero.}
    \label{fig:partial:matching}
\end{figure}

\cite{du2014class} demonstrated that it is possible to perform the described process by minimizing the Pearson divergence (\textrm{Pd}) between $\hat{P}(\textbf{x})$ and $\alpha \hat{P}(\textbf{x}|y(\textbf{x})=\oplus)$, and therefore $\hat{p}$ is estimated as follows:

\begin{equation*}
    \hat{p} = \argmin_{0 \leq \alpha \leq 1}\textrm{Pd}\left(
    \hat{P}(\textbf{x}), \alpha \hat{P}(\textbf{x}|y(\textbf{x})=\oplus)
    \right)
\end{equation*}

The major benefit of this approach (named PE) over EN is that the former drops the need for an intermediate model to accurately estimate the posterior probability, whereas the latter needs a calibrated scorer. However, similarly to EN, PE also overestimates the proportion of positive observations whenever there is overlap between the classes. When there is class overlap, the $\alpha$ derived from minimizing the Pearson Divergence can be so that $\hat{p} \hat{P}(\textbf{x}|y(\textbf{x})=\oplus) > \hat{P}(\textbf{x})$, that is, the Overestimation Area (OA) in Figure~\ref{fig:partial:matching} is greater than zero.

To circumvent the overestimation of PE, \cite{christoffel2016class} introduced pen-L1, which penalizes divergences that cause $\alpha \hat{P}(\textbf{x}|y(\textbf{x})=\oplus) > \hat{P}(\textbf{x})$, therefore limiting the OA.

We introduced a similar method called One Distribution Inside (ODIn) \citep{denisOCQ2018}. In ODIn, the multivariate feature space is mapped onto a univarite space through a one-class scorer, which is induced solely upon $L$ (see more details on one-class scorers in Section~\ref{sec:pat}). After that, the density function is estimated through histograms where the bin boundaries are based on equally spaced percentiles, as illustrated in Figure~\ref{fig:odin:hist}.

\begin{figure}[htb]
\centering
\subfloat[Distribution of training scores.\label{hist_left}]{\includegraphics[scale=1.5]{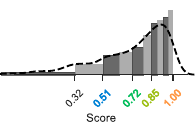}}
\hspace{0.3cm}
\subfloat[Histogram of training scores.\label{hist_right}]{\includegraphics[scale=1.5]{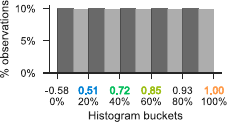}}
\vspace{0.3cm}
\caption{Thresholds for the histogram bins are not uniformly distributed across the scores (\ref{hist_left}) so that each bin is filled with the same proportion of data points (\ref{hist_right}). Source: \cite{denisOCQ2018}.}
\label{fig:odin:hist}
\end{figure}

ODIn applies binary search to maximize an $\alpha$ that causes an Overestimation Area no greater than a given limit $\mathcal{L}$. More specifically, it estimates $\hat{p}$ as follows:

\begin{equation*}
\hat{p} = \mathfrak{s} - \outgrown\left(\mathfrak{s}\right)
\end{equation*}

\noindent where $\outgrown(v)$ is the Overestimation Area when considering $\hat{p} = v$, and $\mathfrak{s}$ is defined as follows:

\begin{equation*}
\mathfrak{s} = \sup_{0 \leq \alpha \leq 1}\left\{\alpha~|~\outgrown\left(\alpha\right) \leq \alpha\mathcal{L}\right\}
\end{equation*}

The limit $\mathcal{L}$ is empirically found through bootstrapping, where $L$ is repeatedly split into two parts and the difference between the areas of the empirical probability density functions of their one-class scores is averaged.
}

\subsection{AlphaMax}

AlphaMax was introduced by \cite{jain2016nonparametric}. In their terminology, $U$ corresponds to the \textit{mixture} sample and $L$ to the \textit{component} sample. The AlphaMax algorithm estimates the maximum proportion of $L$ in $U$.

To better explain the intuition behind AlphaMax, let $U_\oplus \subseteq U$ be a set that contains all positive instances in $U$, and $U_\ominus \subseteq U$ a set that contains all negative instances in $U$. Finally, let $\mathcal{D}(s)$ be the density function of the probability distribution of sample $s$. We know that:

\begin{equation}
\label{eq:alphamax:intui:1}
    \mathcal{D}(U) = (1 - p)\mathcal{D}(U_\ominus) + p \mathcal{D}(U_\oplus)
\end{equation}

Thanks to the assumption of ``selected completely at random'', we also know that $\mathcal{D}(U_\oplus) = \mathcal{D}(L)$. In such a case, we can rewrite Equation~\ref{eq:alphamax:intui:1} as follows:

\begin{equation}
\label{eq:alphamax:intui:2}
    \mathcal{D}(U) =
    (1 - p)\mathcal{D}(U_\ominus)
    + q \mathcal{D}(U_\oplus)
    + r \mathcal{D}(L)
    \hspace{.3cm} \forall \hspace{.1cm} q, r \in [0,1], q + r = p
\end{equation}

In Equation~\ref{eq:alphamax:intui:2}, note that as $r$ increases, $q$ has to proportionally decrease. The objective of AlphaMax is to determine the maximum possible value of $r$, which is $r = p$ when $q = 0$, for which the equation is still valid.



In practice, however, we cannot split $U$ into $U_\ominus$ and $U_\oplus$, since the data is unlabeled. To circumvent this limitation, AlphaMax constructs two density functions, $\tilde{m}$ and $\tilde{c}$, that re-weight the density functions $\hat{m}$ (which estimates the \textit{mixture} $\mathcal{D}(U)$) and $\hat{c}$ (which estimates the \textit{component} $\mathcal{D}(L)$), according to a shared weight vector $\omega$. We emphasize that $\tilde{m}$ specifically counterbalances $\tilde{c}$ by applying it with $1 - \omega$, similarly to the what happens to the component $q\mathcal{D}(U_\oplus)$ of $\mathcal{D}(U)$. For a given $r$, AlphaMax proposes an optimization problem to define $\omega$, given the constraint that $\sum{\omega_i v_i} = r$, where $v_i$ are the weights of $\hat{m}$. For instance, if $\hat{m}$ is estimated using histograms, $v_i$ would be the proportional height of each bin.

The optimization problem tries to maximize a log-likelihood of the mixture (estimation for $\mathcal{D}(U)$) given the weighted participation of the component (estimation for $r\mathcal{D}(L)$). It is stated below:

\begin{equation*}
    ll_r = \max_{\textrm{w.r.t. }\omega}{
        \sum_{\textbf{x} \in U}{
            \log{\tilde{m}(\textbf{x}|\omega)}
        }
        +
        \sum_{\textbf{x} \in L}{
            \log{\tilde{c}(\textbf{x}|\omega)}
        }
    }
\end{equation*}

\begin{figure}[htb]
    \centering
    \includegraphics[width=.4\columnwidth]{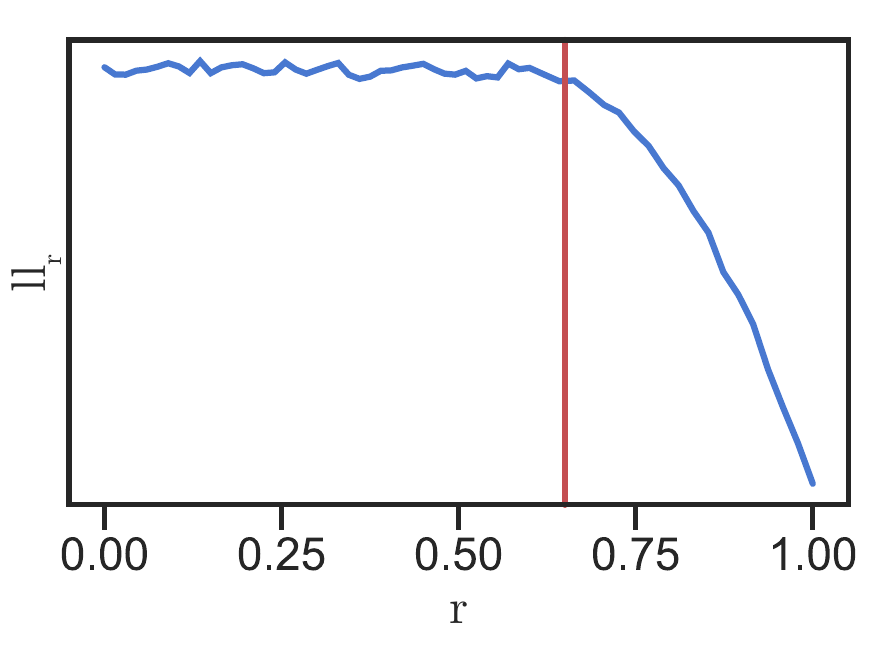}
    \caption{Illustration of the \textit{knee} finding process in AlphaMax. The red vertical line represents a possible contender for the algorithm to report ($r = 0.65$).}
    \label{fig:alphamax:illu}
\end{figure}

Different values of $r$ in the interval $[0,1]$ are applied in the above optimization problem. While $r$ is lower than $p$, it is possible for $\tilde{m}$ to counterbalance $\tilde{c}$, keeping the log-likelihood about the same. However, once the applied $r$ is greater than $p$, the log-likelihood should decrease. AlphaMax returns the value of $r$ that starts the \textit{knee} in the curve of $ll_r$ by $r$, \emph{i.e.}, the value of $r$ that precedes a steep decrease in $ll_r$. Figure~\ref{fig:alphamax:illu} illustrates that process.

An updated version called AlphaMaxN \citep{jain2016estimating} specifically tackles the possibility of the labeled set containing false-positive observations. This setting is out of the scope of this paper. However, we note that in the appendix of \cite{jain2016estimating} there is a mathematically detailed description of the AlphaMax algorithm that is more approachable than the description in its original paper.

At last, we emphasize that solving the optimization problem to define $\omega$ generally is a computationally intensive task that is required several times (one for each value of $r$).

\subsection{KM}

The algorithms that belong to the KM family \citep{ramaswamy2016mixture} have a similar rationale to AlphaMax's. The main difference is that, instead of using log-likelihood to measure the suitability of a possible value for $\hat{p}$ regarding the mixture sample $U$, they use the distances of kernel embeddings. A better comprehension of the algorithm requires deeper understanding of Reproducing Kernel Hilbert Space, which is out of the scope of this paper. 

There are two variants of KM: KM1 and KM2. The difference between the variants is the method to select the gradient threshold, which is used in a similar fashion to the ``knee'' in AlphaMax. KM1 is derived from a theoretical foundation developed by the authors, while KM2 is motivated from empirical evidence.

\subsection{Tree Induction for c Estimation  (TIcE)}
\label{sec:tice}

Tree Induction for $\mathfrak{c}$ Estimation (TIcE) \citep{bekker2018estimating}, as prior PU algorithms, bases its foundation on the assumption of ``selected completely at random''. Observe that Equation~\ref{eq:puq:c2} can be rewritten as follows:

\begin{equation}
\label{eq:tice:puq}
    \mathfrak{c} = \frac{P\left(s(\textbf{x}) = 1\right)}{P\left(y(\textbf{x}) = \oplus\right)}
\end{equation}

From Equation~\ref{eq:tice:puq}, we can derive that a reasonable estimation $\hat{\mathfrak{c}}'$ for $\mathfrak{c}$ is:

\begin{equation}
    \label{eq:tice:ideal}
    \hat{\mathfrak{c}}' = \frac{\left|L\right|}
    {\left|L\right| + \left|U_\oplus\right|}
\end{equation}

\noindent where $U_\oplus \subseteq U$ contains all positive instances in $U$. However, notice that, as $U$ is unlabeled, we \textit{cannot} directly embed $U_\oplus$ in any equation, in practice.

Nonetheless, from Equation~\ref{eq:puq:c} we recall that $P\left(s(\textbf{x}) = 1\right)$ is independent of $\textbf{x}$. In other words, the ratio $\mathfrak{c}$ in Equation~\ref{eq:tice:puq} is constant for any particular region of the feature space.

Consider $\mathcal{S}_\gamma(A) = \{\textbf{x}~|~\textbf{x} \in A, \textrm{ \textbf{x} is within region }\gamma\}$ a function that produces a sub-sample of $A$ that contains all observations that are within the region $\gamma$ of the feature space $\mathcal{X}$. With such a function, we define $\hat{\mathfrak{c}}_\gamma$ as follows:

\begin{equation}
\label{eq:tice:sub}
    \hat{\mathfrak{c}}_\gamma =
    \frac{
        \left|\mathcal{S}_\gamma(L)\right|
    }
    {
        \left|\mathcal{S}_\gamma(L)\right|
        + \left|\mathcal{S}_\gamma(U)\right|
    } =
    \frac{
        \left|\mathcal{S}_\gamma(L)\right|
    }
    {
        \left|\mathcal{S}_\gamma(L)\right|
        + \left|\mathcal{S}_\gamma(U_\oplus)\right|
        + \left|\mathcal{S}_\gamma(U_\ominus)\right|
    }
\end{equation}

\noindent where $U_\ominus \subseteq U$ contains all negative instances in $U$.

Finally, TIcE is interested in finding a region $\gamma$ for which $\hat{\mathfrak{c}}_\gamma$ approximates $\hat{\mathfrak{c}}'$, and therefore $\mathfrak{c}$. To this end, it needs to downplay the influence of $\mathcal{S}_\gamma(U_\ominus)$. Notice that the region $\gamma$ that maximizes $\hat{\mathfrak{c}}_\gamma$ should simultaneously minimize $\left|\mathcal{S}_\gamma(U_\ominus)\right|$, since the remaining of the ratio in Equation~\ref{eq:tice:sub} should approximate the constant value $\mathfrak{c}$ according to the assumption of ``selected completely at random''. Therefore, TIcE proposes the following optimization problem: 

\begin{equation*}
    \hat{\mathfrak{c}} = \max_{\textrm{w.r.t. } \gamma}\{\hat{\mathfrak{c}}_\gamma\}
\end{equation*}

We emphasize that, from the optimization task above, we can derive diverse methods that follow undoubtedly distinct approaches. TIcE, in particular, performs a greedy search by inducing a tree, as we describe next.

In a simplification, to find such a $\gamma$, TIcE uses $U\cup L$ to greedily induce a tree where each node is a sub-region of the feature space within the region defined by its parent node. The node that produces the highest $\hat{\mathfrak{c}}_\gamma$ (given constraints for minimum number of observations) is used to assess one estimation of $\mathfrak{c}$.

We note that although TIcE is introduced as a typical tree-induction algorithm, it is more accurate to describe it as either a greedy search or a biased optimization algorithm, since it uses the estimation assessed by only one node in the tree that may not necessarily be a leaf. Indeed, the algorithm actually intends to locate one region within the feature space. 


{ The greedy tree-induction approach of TIcE has two biases. First, notice that TIcE's greedy search tries to maximize the local estimation of $\hat{\mathfrak{c}}$. This incurs overestimating $\hat{\mathfrak{c}}$. To prevent such an overestimation, several estimations are made via $k$-fold cross validation and the final one is the average of all estimations assessed. In each iteration of the $k$-fold, one fold called tree data is used to drive the greedy search according to the aforementioned criteria. The other $k-1$ folds are concatenated into the estimation data and used to actually estimate $\mathfrak{c}$ after the final region is selected. The estimation data is employed because if, by chance, the tree data has more positive observations in a region than the expected value, such a region would both have a higher chance of being selected and also would derive a higher $\hat{\mathfrak{c}}$ than it should. Additionally, the greed search actually looks for and tries do maximizes estimates of local \textit{lower bounds} of $\hat{c}_\gamma$. To derive such a lower bound, the authors recall that
$|\mathcal{S}_\gamma(L)|\sim \textrm{Binomial}(\mathcal{S}|U_\oplus|,\mathfrak{c})$ and, as such, it is possible to estimate by how much the $\hat{\mathfrak{c}}_\gamma$ can overestimate $\mathfrak{c}_\gamma$ through the one-sided Chebyshev inequality.

}

To better understand the second bias, we describe next how TIcE splits a node. To reduce computational cost, a maximum number of splits is set to avoid excessive computing, and regions of the feature-space are split in order according to a priority-queue so that more promising regions are split first. When TIcE is splitting a region, derived sub-regions are added to the priority-queue. However, the algorithm only adds sub-regions that are created by dividing the feature-space using only \textit{one} feature. More importantly, and contrary to typical tree-induction algorithms for classification, the criterion to choose the feature is based solely on the \textit{one} most promising sub-region derived from a split, despite the possibility of all other resulting sub-regions being unpromising. { In face of such a behavior, one may ask how the algorithm would be affected by the removal or decrease of such bias}. To answer this question, in Section~\ref{sec:exp:eval} we compare TIcE against a proposed extension, Exhaustive TIcE (ExTIcE), described in the next section.

{

Since TIcE tries to maximize lower bounds, it is possible that it ends up underestimating $\mathfrak{c}$. This raises the question on whether it is possible to achieve similar performances with more naive variations of the algorithm.
We realized that a viable alternative to deal with the biases and achieve similar quantification performance is choosing splits completely at random, as we show in Appendix~\ref{appen:ranfoce}.
}

Regarding the computation cost to solve such an optimization problem, the time complexity of TIcE described by \cite{bekker2018estimating} is an overly optimistic $O(mn)$. We better estimate the time complexity of TIcE and the full analysis is presented in Appendix~\ref{appen:tices:complexity}. 

\subsection{Exhaustive TIcE (ExTIcE)}

In this section, we propose Exhaustive TIcE (ExTIcE), an extension of TIcE that aims to lower its search bias.

ExTIcE's main distinction from TIcE is that the former adds all sub-regions created by all features into the priority-queue, while the latter splits the region with only one feature. Despite the name, ExTIcE is not truly Exhaustive. It still sets a hard limit on how many splits can be performed, after which the algorithm is interrupted. We notice that the limit we apply for this paper is the same one applied in TIcE. However, as TIcE always splits the data using only one feature, sub-regions do not share data points and TIcE usually runs out of data before the limit is reached. Conversely, a same data point can be present in several sub-regions for ExTIcE. Additionally, many more sub-regions are added to ExTIcE's priority-queue, even though they will never be split further. For those reasons, ExTIcE is considerably slower than TIcE.

Finally, as is the case with other PUPE approaches described, ExTIcE does not create a reusable model. In the next section, we describe our other proposals, which were originally presented as One-class Quantification methods and are able to induce reusable models.

\subsection{PAT}
\label{sec:pat}

{ In this section we explain and provide theoretical grounding for our one-class quantification method, Passive-Aggressive Threshold (PAT), which we introduced with less detail in previous work \citep{denisOCQ2018}. The main difference from PUPE techniques is that PAT  is based on distribution of scores as a proxy to the distribution of multidimensional data.} We emphasize that, as we do not have training data regarding the negative class, we rely on one-class scorers (OCS). An OCS is a model learned with only positive observations, which outputs, for a previously unseen observation, a numerical value that correlates with the probability of said observation belonging to the positive class. Examples of suitable OCS are One-class SVM \citep{khan2009survey}, Local Outlier Factor \citep{noumir2012simple}, and Isolation Forests \citep{breunig2000lof,liu2008isolation}. In our proposal, we also use the Mahalanobis Distance \citep{mahalanobis1936generalized} as a simple OCS. In this case, the score is the Mahalanobis distance between each observation and the positive data. In all aforementioned algorithms, the score must be either inverted or multiplied by minus one, since they originally are negatively correlated with the probability of belonging to the positive class.

PAT draws inspiration from Adjusted Classify and Count and Conservative Average Quantifier \citep{forman2006quantifying}. As discussed in Section~\ref{sec:def:quant}, ACC depends on accurate estimates for TPR and FPR. However, in many applications we cannot reliably measure either TPR and FPR. This is particularly true for tasks derivative from One-class Quantification, since the distribution of scores of \emph{negative} observations varies from sample to sample.

To offer a better grasp on the intuition behind PAT, observe that the influence of the negative distribution on ACC stems from the fact that the most suitable threshold for \textit{classification} usually cut through the density function of the negative scores, leaving negative scores on both sides of the threshold, as seen in Figure~\ref{fig:bincut}. Although the number of negative observations on the right-hand side of the threshold is expected to be significantly smaller than on the left-hand side, it is still unpredictable whenever the distribution of negative scores changes. 

\begin{figure}[htb]
    \centering
    \includegraphics[width=.8\columnwidth]{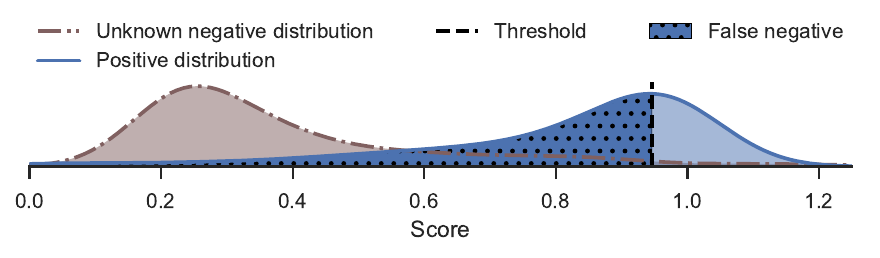}
    \caption{Expected behavior for a conservative threshold: negligible false positive rate and relative small number of true positive observations, in comparison to the total number of positive observations. Source: \cite{denisOCQ2018}}
    \label{fig:pat}
\end{figure}

In PAT, we deliberately choose a very conservative classification threshold that tries to minimize the $\FPR$. In other words, we select a threshold for which we expect very few negative observations to be placed on its right-hand side, as illustrated in Figure~\ref{fig:pat}. With such a conservative threshold, we naively assume that there are no false positive observations. Finally, we extrapolate the total number of expected false negative observations from the number of true positive observations. 

More formally, we set the threshold according to a quantile $\mathfrak{q}$ for the one-class scores of positive observations in a training set. For example, if $\mathfrak{q} = 0.75$, then the threshold is set so that $75\%$ of the training (positive) observations are scored below such a threshold, while $25\%$ are scored above it. Given $\mathfrak{q}$, we estimate $\widehat{\TPR} = 1 - \mathfrak{q}$ and assume $\widehat{\FPR} \approx 0$.

After the threshold is set, we perform ACC as usual: we classify all observations in the test sample of size $|U|$ according to this conservative threshold, count the number of positive instances $N_+$, estimate the positive proportion $\hat{p} = \frac{N_+}{|U|}$, and readjust it as follows:

\begin{equation*}
\hat{p}' = \textit{PAT}(\hat{p}, \mathfrak{q}) = \textit{ACC}(\hat{p}, 1 - \mathfrak{q}, 0) = \min\left\{1, \frac{\hat{p}}{1 - \mathfrak{q}}\right\}
\end{equation*}

\begin{figure}[htb]
\centering
\subfloat[\textit{Dataset Anuran Calls}.\label{fig:pat:var:anuran}]{\includegraphics[width=.45\columnwidth]{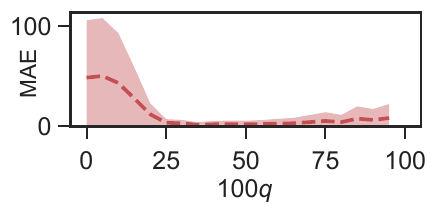}}
\hspace{0.3cm}
\subfloat[\textit{Dataset HRU2}.\label{fig:pat:var:hru2}]{\includegraphics[width=.45\columnwidth]{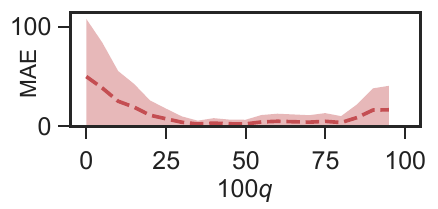}}
\vspace{0.3cm}
\caption{Mean Absolute Error (MAE) of PAT for different values of $\mathfrak{q}$ in two datasets. The shaded area corresponds to two standard deviations. Source: \cite{denisOCQ2018}.}
\label{fig:pat:ar}
\end{figure}

$\mathfrak{q}$ is an important parameter. { Naturally, setting higher values for $\mathfrak{q}$ increases our confidence on our assumption that $\FPR \approx 0$. On the other hand, we can intuitively infer that higher values also increase the methods' error even when the sample only contains positive observations, since we would be extrapolating from fewer data points. To validate this supposition and support a conscious choice regarding $\mathfrak{q}$, we describe relevant statistical characteristics of the method. First, let us make the following assumptions:

\begin{itemize}
    \item[A.1] $\FPR = 0$;
    \item[A.2] the threshold $t_\mathfrak{q}$ set for classification is $\mathfrak{q}$-quantile of the population of scores for the positive class.
\end{itemize}

We recall that A.1 can be formally rewritten as follows, where $S_+$ is the one-class scorer induced upon $L$.

\[
S_+(x) > t_\mathfrak{q} \longrightarrow x \in U_+
\]

Given the previous assumptions, we know that

\[
N_+ \sim \textrm{Binomial}\left(|U_+|,1 - \mathfrak{q}\right)
\]

In this scenario, we can see that the expected value of the error made by PAT is 0:

\begin{align*}
    \EX[p - \hat{p}'] &= p - \frac{\EX[N_+]}{|U|}\frac{1}{1 - \mathfrak{q}}\\
    &= \frac{|U_+|}{|U|} - \frac{|U_+|(1 - \mathfrak{q})}{|U|}\frac{1}{1 - \mathfrak{q}} = 0
\end{align*}

Therefore, if A.1 holds, PAT neither underestimates nor overestimates $p$ \textit{on average}. However, since the variance of the error is not zero, as shown below, the expected value of the absolute error is also not zero:

\begin{align*}
    \Var[p - \hat{p}'] &= \Var[\hat{p}'] = \Var\left[\frac{N_+}{|U|}\frac{1}{1 - \mathfrak{q}}\right]\\
    &= \frac{Var[N_+]}{(|U_+| + |U_-|)^2(1 - \mathfrak{q})^2}
    = \frac{|U_+|\mathfrak{q}(1 - \mathfrak{q})}{(|U_+| + |U_-|)^2(1 - \mathfrak{q})^2}
\end{align*}

We see that, for a given $q$ and $|U|$, increasing $|U_-|$ (and therefore equally decreasing $|U_+|$) only contributes to decreasing the variance of $\hat{p}$. In other words, the variance is bigger for samples that contain only positive observations. Finally, we see that it is also bigger for smaller samples.

The expected absolute error $\EX[|p - \hat{p}|]$ (EAE for short) shall inherit the behavior of $\Var(\hat{p})$. It can be obtained through the Law of the unconscious statistician, as follows:

\begin{align*}
    \EX[|p - \hat{p}'|] &=
    \sum_{i=0}^{|U_+|}\left|\frac{|U_+|(1 - \mathfrak{q}) - i}{(|U_+| + |U_-|)(1 - \mathfrak{q})}\right| P(N_+ = i)\\
    &= \sum_{i=0}^{|U_+|}\left|\frac{|U_+|(1 - \mathfrak{q}) - i}{(|U_+| + |U_-|)(1 - \mathfrak{q})}\right|
    \binom{|U_+|}{i}
    \mathfrak{q}^{|U_+| - i}(1 - \mathfrak{q})^{i}
\end{align*}

The variance of the absolute error can be obtained by noting that

\begin{align*}
\Var[|p - \hat{p}'|] &= \EX[|p - \hat{p}'|^2] - (\EX[|p - \hat{p}'|])^2\\
&= \EX[(p - \hat{p}')^2] - (\EX[|p - \hat{p}'|])^2\\
&= \Var[p - \hat{p}'] + (\EX[p - \hat{p}'])^2 - (\EX[|p - \hat{p}'|])^2\\
&= \Var[\hat{p}'] - (\EX[|p - \hat{p}'|])^2
\end{align*}

\noindent which only uses terms we previously calculated.


We remind that the Mean Absolute Error (MAE), which we employ to measure the performance in our experimental evaluation, estimates the expected absolute error.

In Figure~\ref{fig:expected:abs:err} we show the expected absolute error for a varying number of test sample sizes (from $50$ to $2{,}000$). For this figure, we consider that the test sample would have only positive observations, since this situation incurs the highest EAE, and therefore is the worst case.

\begin{figure}[htb]
    \centering
    \includegraphics[width=\columnwidth]{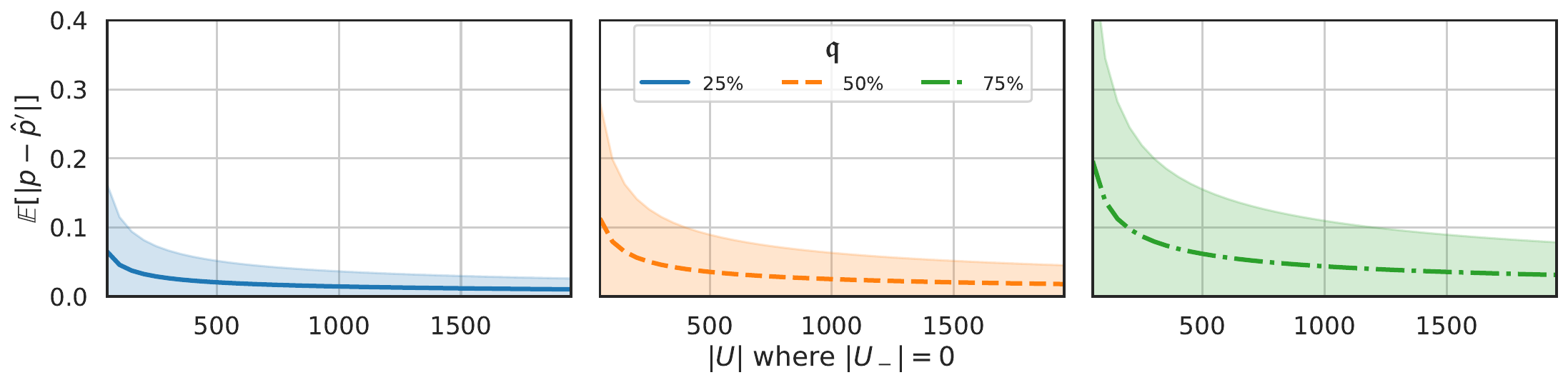}
    \caption{Expected absolute error for PAT. Sample contains only positive observations and sample size varies from $50$ to $2{,}000$. The shaded area represents two standard deviations.}
    \label{fig:expected:abs:err}
\end{figure}

We notice two important characteristics of the method. First, the EAE decays exponentially as we increase the sample size. Therefore, quantifying larger samples yields more confidence on the estimates. This result is unsurprising since, generally speaking, larger samples yield better estimations for characteristics of a population. The second characteristic is that smaller values of $\mathfrak{q}$ incur smaller EAE.

With the previous results in mind, we have two opposing forces directing our choice regarding $\mathfrak{q}$. On the one hand, smaller values lead to smaller EAE. On the other hand, it becomes more difficult to be lean on A.1 if $\mathfrak{q}$ is too small. We recall that, after all, how small $\mathfrak{q}$ can be depends on two factors: which one-class scorer we adopted for our data and the behaviour of scores of the negative class, which we cannot predict precisely.

We further notice that, if the classes are completely separable in the feature space, a good enough one-class scorer would allow us to choose $\mathfrak{q}=0$, since $\mathfrak{q}$ is a quantile of the distribution of scores for only \textit{positive} data points. Choosing $\mathfrak{q} > 0$ implicitly sets an expectation that the classes overlap. For instance, a $\mathfrak{q}=0.25$ means we expect that up to $25\%$ of the lower scored data points would be compromised by overlap between classes. In this sense, even $0.25$ is a considerably big number.

In previous work \citep{denisOCQ2018}, we experimentally showed a broad range of possible values for $\mathfrak{q}$ lead to similar quantification errors in practice, as illustrated in Figure~\ref{fig:pat:ar}.

} 

Therefore, in an attempt to achieve a reasonable compromise between simultaneously addressing the high EAE, caused by bigger values of $\mathfrak{q}$, and the higher risk of A.1 not holding for small values of $\mathfrak{q}$, we proceeded as follows: instead of picking a single value to be used in our experiments, we adopted a strategy similar to Median Sweep \citep{forman2006quantifying}. In this case, we apply PAT with $\mathfrak{q}$ from $0.25$ to $0.75$ with increases of $0.01$ and consider the median of the estimates.

Regarding the time complexity of PAT, since we can reuse the scorer model multiple times, we split the analysis into two stages: training and test.

For the training stage, consider $\mathcal{V}(n)$ to be the time complexity of training a scorer with $n$ observations, and $\nu$ the time complexity of scoring one single observation. Suppose that we apply $k$-fold cross validation to obtain the positive scores, with which we model the density function to identify the $t$ thresholds associated with different values of $\mathfrak{q}$. In this case, the complexity to train PAT is the complexity to obtain the scores and sort them in order to identify the thresholds (either linearly or through binary search):

\begin{equation*}
O\left(
k\mathcal{V}\left(\frac{k-1}{k}|L|\right) + |L|\nu + |L|\log|L|
\right)
\end{equation*}

For test, we can take different approaches if we are using multiple thresholds or only one. If we are using only one threshold, then, after scoring all test observations, we can linearly count how many are below the threshold, totalling a time complexity of $O(|U|\nu + |U|) = O(|U|\nu)$. However, if we are using multiple thresholds, we can sort the scores and iterate over a pre-sorted list of thresholds to count how many observations are below each threshold with binary search. In this case, the time complexity is $O(|U|\nu + |U|log|U|)$.

%% file: sections/expsetup.tex
\section{Experimental Setup}
\label{sec:exp:setup}

In this section, we explain the experimental setup and datasets used in our empirical evaluation. 
In the general setting, for each dataset, we varied the true positive ratio, \emph{i.e.}, the proportion of the positive class in the unlabeled (test) sample, from  0\% to 100\% with increments of 10\%. { For a given positive ratio, we split the data into five folds, and we perform five evaluations for the quantification task. In each evaluation, one fold has its negative examples excluded and all positive examples are labeled, while the other four folds are concatenated without labels.}

Due to the slowness of some of the algorithms tested, training data size was limited to 500 observations and test data size was limited to 2,000 observations. Final results are reported as Mean Absolute Error (MAE), which is the average of the absolute difference between the predicted positive ratio and the true positive ratio.

Finally, we raise attention to the fact that if we employ a { baseline} quantifier that always predict $\hat{p} = 0.5$, and the actual $p$ is uniformly distributed within the range $[0,1]$ (as in our experiments, { since we linearly iterate over this range}), then the MAE obtained in a large enough number of test samples converges to $0.25$. 

{
To better illustrate this, let $X \sim \textrm{Uniform}(0, 1)$ be a random variable that represents the actual proportion of positive examples. Since such a baseline always estimate it to be $0.5$, the expected absolute error (estimated by MAE) is:

\begin{align*}
    \EX[|X-0.5|] &= \int_0^1 |x - 0.5| dx\\
    &= \int_0^{0.5}(0.5 - x )dx + \int_{0.5}^1(x - 0.5)dx = 0.25
\end{align*}

}

This fact indicates that the maximum error we should consider as acceptable in our setting is $25\%$.

All code is avaliable in our supplemental material website \citep{supmatweb}. Next, we describe the particularities of each experiment.

\subsection{Experiment \#1}

In this experiment, the existence of negative sub-classes is disregarded. The size of the test sample is the mininum between the number of available positive and negative candidates, limited to 2,000. The number of repetitions is five. This experiment is designed to be easy to reproduce and compare, although supports only a superficial analysis of performance. Our objective with this experiment is to provide a similar setup with those in current literature, and to provide a general analysis of quantification performance. 

\subsection{Experiment \#2}
\label{sec:exp:2}

In this experiment, the existence of negative sub-classes is acknowledged. For each test sample, the proportion of the negative sub-classes is randomized and the sample is drawn accordingly. The size of the test sample is the largest that make the previously set proportion viable, limited to a maximum of 500 observations. To obtain greater variability in the test samples, giving the random proportion of sub-classes, the number of repetitions is 30. With this setting, we aim to produce experimental results that better suit our assumption that the negative class varies from sample to sample. Experiment \#2 is the same as the one we employed in our previous work to measure the performance of PAT \citep{denisOCQ2018}. Next, we describe a relevant limitation of this setting and how we overcome it.

\subsection{Experiment \#3}
\label{sec:exp:3}

The uniform randomization of the proportions of negative sub-classes, in Experiment \#2, has an adverse effect. While the MAE for each individual sub-class proportion is informative for the expected performance for said proportion, the experimental MAE when averaging all variations of sub-class proportions is bound to converge to the same MAE that would be obtained with balanced test samples, that is, test samples whose every single sub-class has the same number of observations.

However, in real world applications, we do not assume that all classes will appear with the same proportion. On the contrary, we assume that the proportion of the sub-classes vary and is unknown beforehand. To better evaluate the methods in this situation, we propose Experiment \#3. In this experiment, we map the original dataset onto several datasets, one \textit{for each} negative sub-class, containing data points of a single negative sub-class and all positive data points. Each dataset is evaluated individually. The size of the test samples is the minimum between the number of available positive and negative candidates, limited to 2,000. Finally, we evaluate:

\begin{description}
    \item[Experiment \#3-a --\textbf{ Median}] half of the negative classes produced MAE lower or equal than the one reported in this experiment;
    \item[Experiment \#3-b --\textbf{ 75-percentile}] three quarters of the negative classes produced MAE lower or equal than the one reported in this experiment;
    \item[Experiment \#3-c --\textbf{ Worst case}] the result obtained by the single negative class that produced the greatest MAE.
\end{description}

\subsection{Experiment \#4}
\label{sec:exp:4}

The aim of this experiment is to compare execution time of different algorithms.

Due to the slowness of some of the algorithms evaluated, the previous experiments were executed in parallel in a variety of hardware across multiple days. To measure the time consumed by each algorithm in a comparable manner, we performed a diminished version of Experiment \#1 that was executed in a single machine. The differences are: 5-fold cross validation was interrupted after the evaluation of the first fold, and the experimental setup was evaluated only once instead of repeating for five times.

We highlight that the time necessary to quantify each test sample was measured independently, and summed at the end, to avoid measuring time spent with the preparation of the samples.

\subsection{Algorithms}
\label{sec:exp:algs}

We evaluated the performance of seven algorithms: EN, PE, KM1, KM2, TIcE, ExTIcE, and PAT. All methods were merged into a unified test framework, publicly available as supplemental material \citep{supmatweb}.

Given the algorithm's simplicity, we used our own implementation for EN. \cite{elkan2008} employed Support Vector Machine calibrated with Platt scaling as a base classifier for EN. In this article, we adopted the same method to keep compatibility between experiments.

EN relies on SVM. We used scikit-learn's implementation \citep{scikit-learn} with all parameters set to default, excepting gamma. Gamma is an important parameter that usually is set to either ``auto'' or ``scale''. This parameter caused severe differences in the results for some datasets. For this reason, we report results for both settings, where EN\textsubscript{a} refers to the situation where gamma is set to ``auto'', and EN\textsubscript{s}'' refers to the situation where gamma is set to ``scale''.

The code of PE, used in our experiments, was a direct translation to Python 3 from the original code, in Matlab, provided by \cite{du2014class}\footnote{\url{http://www.mcduplessis.com/index.php/}}.

Code for pen-L1 is not available in the author's website. However, comparisons are possible due to transitivity and analysis of previous work \citep{bekker2018estimating}. In other words, we assume that if one algorithm $A$ performs better than $B$ in our experiments, and $B$ performs better than $C$ in the existing literature, $A$ performs better than $C$.

We reached out to the AlphaMax's authors and they attentively provided us with code and instructions to use AlphaMax in our experiments. Unfortunately, a fair use of the program provided would require several manual interventions. Given the volume of the experiments in our setup, making such interventions would be unfeasible and unfair with the other contenders. Alternatively, results for AlphaMax are provided in previous work \citep{ramaswamy2016mixture,bekker2018estimating}, so that it is possible to draw some conclusions by assuming transitivity.

For KM1 and KM2, we used code provided by their original authors \citep{ramaswamy2016mixture}\footnote{\url{http://web.eecs.umich.edu/~cscott/code.html}}. The code of KM1 and KM2 is a single script that produces results for both variants, since they share the significant part of the computation required to evaluate a sample. For this reason, in Experiment \#4, time spent for both algorithms is aggregated into a single column, KM.

Although we previously tested PAT with different scorers \citep{denisOCQ2018}, in our analysis, we keep only the results for PAT with Mahalanobis distance (PAT\textsubscript{M}). We chose PAT\textsubscript{M} to be a representative of PAT in our comparisons against PU techniques since Mahalanobis Distance is the simplest scorer among the ones cited in this work and does not require any parameter, and having one single version of PAT simplifies our analysis. Another important difference regarding our previous usage of PAT is that, here, we vary the parameter $q$ from 25\% to 75\% with increments of one and report the median of all predictions, instead of fixing the parameter to a single value.

As PAT\textsubscript{M} is the only algorithm tested that produces a model that can be used for several test samples, in Experiment \#4 we \textit{additionally} report the time spent by PAT\textsubscript{M} to only quantify the data, while disregarding the time spent with training.

The implementation of TIcE provided by \cite{bekker2018estimating} only supports categorical features after binarization. Furthermore, numerical features should be in the range $[0,1]$. Yet, when a numerical feature is selected to split a node, only four sub-regions are created for the ranges $[0,0.25]$, $[0.25,0.5]$, $[0.5,0.75]$ and $[0.75,1]$. Since this implementation handles numerical data too simplistically and we only use numerical datasets, we developed our own implementation for TIcE. For each split, we divide the region into two sub-regions with roughly the same number of observations: one with all observations that are below or equal the median of the splitting feature, and the other with the remaining. { We adopt numpy's\footnote{\url{https://numpy.org/}} method to compute the median, which applies introselect and has linear performance for sample sizes with less than $2{,}000$ data points, such as the ones we use. We bench-marked its performance and show it in Figure~\ref{fig:numpy:median}. Therefore, the way we compute the median does not increase the time complexity of the algorithm.

\begin{figure}[htb]
    \centering
    \includegraphics[width=0.5\columnwidth]{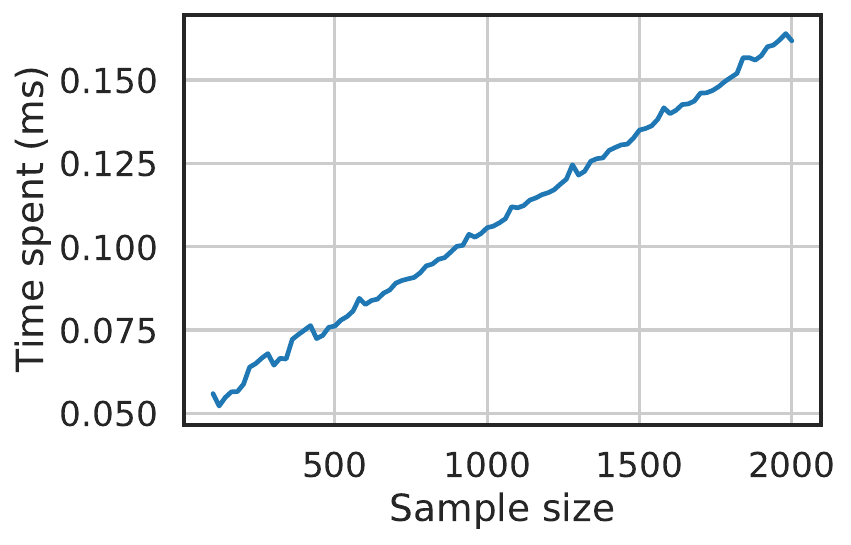}
    \caption{Time spent to compute the median using numpy's method. Each data point is the median over $10$ iterations, where, for each iteration, the accumulated time is measured for $200$ samples and later divided by $200$.}
    \label{fig:numpy:median}
\end{figure}

}

Throughout experiments \#1 to \#3, we additionally compare all algorithms against a hypothetical classify and count approach that uses the Mahalanobis distance as scorer and the best possible threshold for classification. { This algorithm is called Best Fixed Threshold \citep{denisOCQ2018} with Mahalanobis, or BFT\textsubscript{M} for short.} To choose the threshold, we evaluate several thresholds based on the percentiles of the positive training data (from 0 to 100, with increments of 1). For each dataset, we evaluate which threshold generated the lowest MAE on the test samples and report such a result. We emphasize that, regardless of the average performance obtained by BFT\textsubscript{M}, it still is affected by the systemic error explained in Section~\ref{sec:def:quant}.

\subsection{Datasets}
\label{sec:datasets}

In our experiments, we used 10 datasets. Nine are directly derived from real data, and one is generated by a Bayesian network. To maintain consistency, for each dataset, the positive class is the same as in our previous work \citep{denisOCQ2018}, where they were chosen arbitrarily. These datasets were chosen due to all being publicly available and having a large enough number of observation points, for each class, so that we can proceed with the data hungry experimental setup described in Section~\ref{sec:exp:setup}. Each dataset is detailed below:

\begin{description}

\item[I\underline{n}sects v2] sensor data regarding the flight of $18$ classes of insects. A class of insect is determined by sex and species. The observations are described by $10$ features extracted from a time series obtained from a single sensor. No environmental feature is included. All data was collected within a temperature range from $27$ (included) to $30$ (included) degree Celsius. The number of observations per class was limited to $10{,}000$ observations (achieved by seven classes). The class with the least number of observations has $259$. The total number of records is $83{,}550$, and the positive class is female \textit{Aedes aegypti} with $10{,}000$ observations;

\item[\underline{I}nsects] contains information about the flight of $14$ species of insects. As some are discriminated further by sex, the dataset has $18$ classes. The positive class is female \textit{Aedes aegypti}. The data has $166{,}880$ records represented by $27$ features. We find this dataset to be heavily biased regarding the environmental feature temperature. This dataset was kept in our evaluations only to maintain consistency with our previous work \citep{denisOCQ2018};    

\item[\underline{A}rabic Digit] contains $8{,}800$ entries described by $26$ features for the human speech of Arabic digits. There are $10$ classes, and the target class is the digit $0$. This version sets a fixed number of features for every record \citep{hammami2010improved,Lichman:2013};

\item[\underline{B}NG (Japanese Vowels)] Bayesian network generated benchmark dataset with speech data regarding Japanese Vowels. There are $1{,}000{,}000$ entries, represented by $12$ features, for $9$ speakers. The speaker $\#1$ is the class of interest \citep{OpenML2013};

\item[Anuran \underline{C}alls (MFCCs)] contains $22$ features to represent the sound produced by different species of Anurans (frogs). As the data size is restricted, we only considered the two biggest families of frogs as the classes of the data, ending up with $6{,}585$ entries. The positive class is the \textit{Hylidae} family, and the negative class is the \textit{Leptodactylidae} family \citep{diaz2012compressive,Lichman:2013};

\item[\underline{H}andwritten] contains $63$ features that represent the handwritten lowercase letters \textit{q}, \textit{p} and \textit{g}. The data has $6{,}014$ entries and the chosen positive class is the letter \textit{q} \citep{dmr2018unsupervised};

\item[\underline{L}etter] describes the appearance of the $26$ uppercase letters of the alphabet on a black and white display with $16$ features. It contains $20{,}000$ entries and the class of interest is the letter \textit{W} \citep{frey1991letter,OpenML2013};

\item[\underline{P}en-Based Recognition of Handwritten Digits] handwritten digits represented by $16$ features. The digit $5$ is the target class. There are $10{,}992$ entries \citep{alimoglu1996combining,Lichman:2013};

\item[H\underline{R}U2] Pulsar candidates collected during the HTRU survey, where pulsars are a type of star. It contains two classes, Pulsar (positive) and not-Pulsar (negative), across $17{,}898$ entries described by $63$ features \citep{lyon2016fifty,Lichman:2013};

\item[\underline{W}ine Quality] contains $11$ features that describe two types of wine (white and red). The quality information was disregarded, and the target class is red wine. The dataset contains $6{,}497$ entries \citep{cortez2009modeling,Lichman:2013}.
\end{description}

KM1 and KM2 presented a runtime error while processing dataset H (Handwritten). For that reason, the performance of these algorithms is not present in any tables for this dataset.

%% file: sections/expeval.tex
\section{Experimental Evaluation}
\label{sec:exp:eval}

In this section, we display and analyze the results we obtained with the experiments explained in the previous section. For all experiments, we present the average rank and, from completeness, a critical difference plot for the Nemenyi test with $\alpha=0.1$. This test is intended as a simple way of comparing all algorithms in one go. However, we observe the limitations of this test as it only takes the ranks into account and is conservative with the amount of data we have. In some cases, the difference between some results are glaring, even in different orders of magnitude, and the test fails to recognize the superiority of some approaches. We make particular observations for such cases and perform pair-wise comparisons via Wilcoxon signed-rank test, when relevant.

\begin{figure}[htb]
    \centering
    \includegraphics[width=.75\columnwidth]{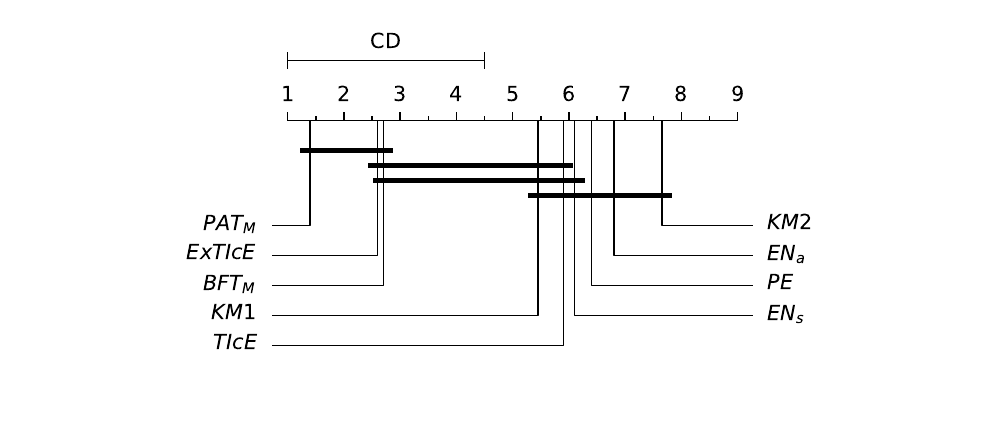}
    \caption{Nemenyi test for Experiment \#1 with $\alpha=0.1$. Methods within the \textbf{c}ritical \textbf{d}ifference (CD) are connected by horizontal line and are not significantly different.}
    \label{fig:nemenyi:basic}
\end{figure}

Table~\ref{tab:err:xp:basic} summarizes our results for Experiment \#1, and Figure~\ref{fig:nemenyi:basic} shows the corresponding critical difference plot. PAT\textsubscript{M}, our proposal, outperformed all PU approaches in 9 out of 10 datasets. It underperformed (within one standard deviation) KM1 and ExTIcE only in the dataset Insects v2, in which PAT\textsubscript{M} ranked third. We observe that, as expected, PAT\textsubscript{M} outperformed BFT\textsubscript{M} in most cases. Although BFT\textsubscript{M} is overly optimistic since the threshold is chosen based on the final results, it still undergoes CC's systemic error explained in Section~\ref{sec:definitions}. 

\begin{table}[htb]
\centering
\scriptsize
\begin{tabular}{r|cccccccc|c}
\hline
Data & EN\textsubscript{a} & EN\textsubscript{s} & PE & KM1 & KM2 & TIcE & ExTIcE & PAT\textsubscript{M} & BFT\textsubscript{M} \\
\hline
\rowcolor{gray!25}
 & 37.02 & 14.60 & 15.94 & 7.98 & 17.06 & 16.59 & \textbf{7.72} & 8.92 & 8.45 \\
\rowcolor{gray!25}
\multirow{-2}{*}{N} & \tiny(22.99) & \tiny(6.42) & \tiny(3.94) & \tiny(5.10) & \tiny(4.46) & \tiny(2.73) & \textbf{\tiny(2.78)} & \tiny(2.43) & \tiny(1.49) \\
 & 26.92 & 25.62 & 20.55 & 12.91 & 14.63 & 21.72 & 12.78 & \textbf{7.34} & 10.28 \\
\multirow{-2}{*}{I} & \tiny(2.81) & \tiny(4.85) & \tiny(1.51) & \tiny(4.76) & \tiny(4.41) & \tiny(3.56) & \tiny(3.55) & \textbf{\tiny(2.48)} & \tiny(1.71) \\
\rowcolor{gray!25}
 & 10.78 & 11.14 & 13.34 & 25.43 & 26.39 & 20.49 & 13.12 & \textbf{3.87} & 6.45 \\
\rowcolor{gray!25}
\multirow{-2}{*}{A} & \tiny(5.47) & \tiny(4.88) & \tiny(6.15) & \tiny(10.25) & \tiny(6.92) & \tiny(4.77) & \tiny(3.40) & \textbf{\tiny(2.79)} & \tiny(2.40) \\
 & 13.98 & 10.94 & 16.66 & 13.37 & 18.48 & 17.25 & 8.85 & \textbf{5.76} & 10.04 \\
\multirow{-2}{*}{B} & \tiny(3.79) & \tiny(2.93) & \tiny(2.35) & \tiny(8.25) & \tiny(5.08) & \tiny(2.76) & \tiny(3.06) & \textbf{\tiny(2.37)} & \tiny(1.81) \\
\rowcolor{gray!25}
 & 14.86 & 12.64 & 11.79 & 12.02 & 15.25 & 11.32 & 4.26 & \textbf{2.03} & 10.85 \\
\rowcolor{gray!25}
\multirow{-2}{*}{C} & \tiny(4.82) & \tiny(4.63) & \tiny(2.09) & \tiny(5.71) & \tiny(3.92) & \tiny(2.07) & \tiny(2.30) & \textbf{\tiny(1.63)} & \tiny(3.77) \\
 & 8.03 & 49.02 & 12.57 & \multirow{2}{*}{--} & \multirow{2}{*}{--} & 11.37 & 5.68 & \textbf{3.66} & 4.76 \\
\multirow{-2}{*}{H} & \tiny(3.34) & \tiny(1.90) & \tiny(3.08) &  &  & \tiny(2.04) & \tiny(2.61) & \textbf{\tiny(4.69)} & \tiny(3.32) \\
\rowcolor{gray!25}
 & 8.97 & 10.42 & 12.49 & 16.02 & 19.57 & 10.23 & 5.84 & 3.18 & \textbf{2.99} \\
\rowcolor{gray!25}
\multirow{-2}{*}{L} & \tiny(6.95) & \tiny(6.24) & \tiny(5.71) & \tiny(8.22) & \tiny(5.64) & \tiny(3.39) & \tiny(3.19) & \tiny(2.33) & \textbf{\tiny(1.66)} \\
 & 50.04 & 10.58 & 10.75 & 12.86 & 20.80 & 13.45 & 6.81 & \textbf{2.57} & 2.71 \\
\multirow{-2}{*}{P} & \tiny(3.59) & \tiny(5.56) & \tiny(3.67) & \tiny(6.94) & \tiny(4.27) & \tiny(4.16) & \tiny(3.31) & \textbf{\tiny(1.98)} & \tiny(1.47) \\
\rowcolor{gray!25}
 & 24.19 & 14.46 & 15.44 & 6.46 & 10.47 & 7.86 & 4.17 & \textbf{2.64} & 11.76 \\
\rowcolor{gray!25}
\multirow{-2}{*}{R} & \tiny(11.21) & \tiny(4.73) & \tiny(2.94) & \tiny(4.77) & \tiny(3.07) & \tiny(2.13) & \tiny(2.57) & \textbf{\tiny(2.09)} & \tiny(3.42) \\
 & 12.12 & 21.98 & 16.20 & 5.45 & 8.77 & 8.56 & 3.23 & \textbf{2.23} & 4.94 \\
\multirow{-2}{*}{W} & \tiny(5.31) & \tiny(4.86) & \tiny(2.37) & \tiny(3.91) & \tiny(4.28) & \tiny(1.97) & \tiny(2.46) & \textbf{\tiny(1.87)} & \tiny(2.18) \\
\hline
$\overline{\textrm{rank}}$ & 6.8 & 6.1 & 6.4 & 5.5 & 7.7 & 5.9 & 2.6 & 1.4 & 2.7 \\
\hline
\end{tabular}
\caption{Mean absolute error (standard deviation in parentheses), in percentages, for experiment \#1.}
\label{tab:err:xp:basic}
\end{table}

Also as expected, ExTIcE outperformed TIcE in every dataset, since ExTIcE removes a search constraint from TIcE. However, more noteworthy is the fact that ExTIcE performed better than all PULearning approaches in all datasets, although a direct comparison against BFT\textsubscript{M} is inconclusive (p-value is one for Wilcoxon Rank-Sum test).

Regarding Figure~\ref{fig:nemenyi:basic}, we note that although PAT\textsubscript{M} did not differ significantly from ExTIcE, the test only evaluated the average rank of the algorithms. Directly comparing PAT\textsubscript{M} against ExTIcE with Wilcoxon Rank-Sum test results in a p-value of $0.2$.


The results from Experiment \#2 were unremarkably similar to the ones from Experiment \#1. This is due to the fact that the majority of the datasets used in our experiments are already fairly balanced (regarding the negative sub-classes). For this reason, we do not further analyze such results. They are displayed in Appendix~\ref{appen:exp:2}.

\begin{figure}[htb]
    \centering
    \includegraphics[width=.75\columnwidth]{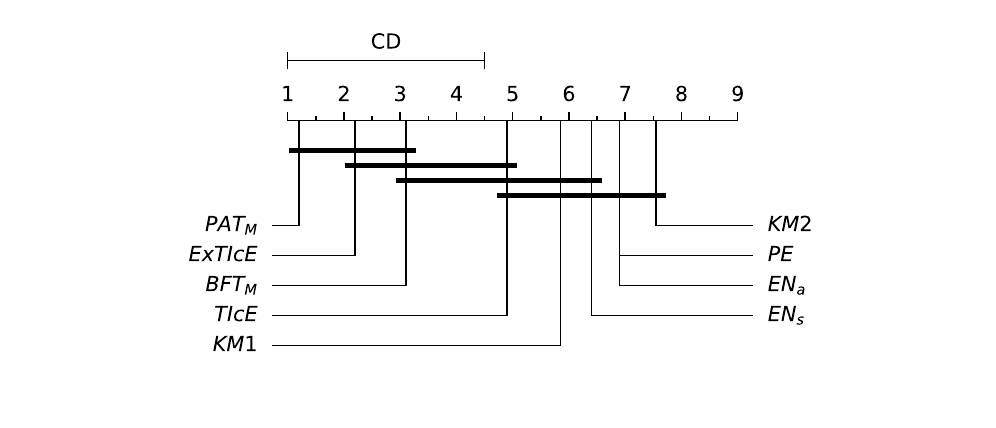}
    \caption{Nemenyi test for Experiment \#3-a with $\alpha=0.1$. Methods within the \textbf{c}ritical \textbf{d}ifference (CD) are connected by horizontal line and are not significantly different.}
    \label{fig:nemenyi:median}
\end{figure}

Table~\ref{tab:err:xp:median} and Figure~\ref{fig:nemenyi:median} present the results for Experiment \#3-a. As a recap, the results in the table indicates that, for half the classes, the MAE obtained is lower or equal than the value shown. While the rankings are mostly unchanged from Experiments \#1 and \#2, we observe that for some datasets, especially N, I, and B, the MAE obtained by both ExTIcE and PAT\textsubscript{M} are below half of those obtained in the previous experiments. This is evidence of a great disparity in the separability between different sub-classes and the positive class. We can therefore expect that the PAT's low errors in Experiment \#3-a should be compensated by larger errors as we investigate more difficult sub-classes, in Experiments \#3-b and \#3-c.

\begin{table}[htb]
\centering
\scriptsize
\begin{tabular}{r|cccccccc|c}
\hline
Data & EN\textsubscript{a} & EN\textsubscript{s} & PE & KM1 & KM2 & TIcE & ExTIcE & PAT\textsubscript{M} & BFT\textsubscript{M} \\
\hline
\rowcolor{gray!25}
 & 32.82 & 8.50 & 15.98 & 11.18 & 14.48 & 7.93 & 3.53 & \textbf{2.09} & 5.29 \\
\rowcolor{gray!25}
\multirow{-2}{*}{N} & \tiny(12.49) & \tiny(5.54) & \tiny(1.89) & \tiny(5.19) & \tiny(2.78) & \tiny(1.27) & \tiny(1.81) & \textbf{\tiny(1.75)} & \tiny(1.70) \\
 & 20.05 & 20.36 & 16.17 & 7.45 & 9.12 & 10.92 & 5.23 & \textbf{2.16} & 7.50 \\
\multirow{-2}{*}{I} & \tiny(3.02) & \tiny(2.97) & \tiny(1.32) & \tiny(3.54) & \tiny(2.81) & \tiny(1.42) & \tiny(2.35) & \textbf{\tiny(1.67)} & \tiny(2.04) \\
\rowcolor{gray!25}
 & 8.54 & 9.65 & 13.69 & 14.30 & 19.33 & 10.81 & 6.44 & \textbf{3.73} & 6.79 \\
\rowcolor{gray!25}
\multirow{-2}{*}{A} & \tiny(5.80) & \tiny(6.12) & \tiny(5.09) & \tiny(7.72) & \tiny(5.38) & \tiny(3.59) & \tiny(2.28) & \textbf{\tiny(2.69)} & \tiny(2.53) \\
 & 10.26 & 7.72 & 13.51 & 10.23 & 14.81 & 11.22 & 4.89 & \textbf{3.21} & 9.07 \\
\multirow{-2}{*}{B} & \tiny(3.18) & \tiny(2.93) & \tiny(1.79) & \tiny(4.53) & \tiny(4.42) & \tiny(2.12) & \tiny(2.17) & \textbf{\tiny(1.99)} & \tiny(2.01) \\
\rowcolor{gray!25}
 & 14.86 & 12.64 & 11.79 & 12.02 & 15.25 & 11.32 & 4.26 & \textbf{2.03} & 10.74 \\
\rowcolor{gray!25}
\multirow{-2}{*}{C} & \tiny(4.82) & \tiny(4.63) & \tiny(2.09) & \tiny(5.71) & \tiny(3.92) & \tiny(2.07) & \tiny(2.30) & \textbf{\tiny(1.63)} & \tiny(3.63) \\
 & 7.93 & 49.57 & 11.87 & \multirow{2}{*}{--} & \multirow{2}{*}{--} & 11.12 & 5.49 & \textbf{3.34} & 6.48 \\
\multirow{-2}{*}{H} & \tiny(3.28) & \tiny(1.05) & \tiny(2.46) &  &  & \tiny(1.97) & \tiny(2.34) & \textbf{\tiny(3.76)} & \tiny(3.76) \\
\rowcolor{gray!25}
 & 8.03 & 10.26 & 11.93 & 12.09 & 18.20 & 6.53 & 3.73 & 3.07 & \textbf{2.55} \\
\rowcolor{gray!25}
\multirow{-2}{*}{L} & \tiny(5.38) & \tiny(5.69) & \tiny(4.42) & \tiny(6.81) & \tiny(5.31) & \tiny(2.86) & \tiny(2.53) & \tiny(2.38) & \textbf{\tiny(1.17)} \\
 & 49.84 & 11.92 & 10.09 & 12.29 & 17.04 & 7.84 & 4.62 & 2.58 & \textbf{2.54} \\
\multirow{-2}{*}{P} & \tiny(1.01) & \tiny(5.63) & \tiny(3.39) & \tiny(5.18) & \tiny(3.80) & \tiny(3.18) & \tiny(2.68) & \tiny(2.11) & \textbf{\tiny(0.94)} \\
\rowcolor{gray!25}
 & 24.19 & 14.46 & 15.44 & 6.46 & 10.47 & 7.86 & 4.17 & \textbf{2.64} & 11.49 \\
\rowcolor{gray!25}
\multirow{-2}{*}{R} & \tiny(11.21) & \tiny(4.73) & \tiny(2.94) & \tiny(4.77) & \tiny(3.07) & \tiny(2.13) & \tiny(2.57) & \textbf{\tiny(2.09)} & \tiny(3.29) \\
 & 12.12 & 21.98 & 16.20 & 5.45 & 8.77 & 8.56 & 3.23 & \textbf{2.23} & 4.95 \\
\multirow{-2}{*}{W} & \tiny(5.31) & \tiny(4.86) & \tiny(2.37) & \tiny(3.91) & \tiny(4.28) & \tiny(1.97) & \tiny(2.46) & \textbf{\tiny(1.87)} & \tiny(2.21) \\
\hline
$\overline{\textrm{rank}}$ & 6.9 & 6.4 & 6.9 & 5.8 & 7.5 & 4.9 & 2.2 & 1.2 & 3.1 \\
\hline                    
\end{tabular}
\caption{Mean absolute error (standard deviation in parentheses), in percentages, for Experiment \#3-a (median of performance among negative sub-classes). Results for datasets C, R and W are repetitions from Table~\ref{tab:err:xp:basic}, since they contain only one negative sub-class.}
\label{tab:err:xp:median}
\end{table}

\begin{table}[htb]
\centering
\scriptsize
\begin{tabular}{r|cccccccc|c}
\hline
Data & EN\textsubscript{a} & EN\textsubscript{s} & PE & KM1 & KM2 & TIcE & ExTIcE & PAT\textsubscript{M} & BFT\textsubscript{M} \\
\hline
\rowcolor{gray!25}
 & 40.02 & 10.05 & 27.04 & 12.25 & 18.01 & 8.62 & 5.05 & \textbf{3.17} & 7.78 \\
\rowcolor{gray!25}
\multirow{-2}{*}{N} & \tiny(21.10) & \tiny(5.96) & \tiny(1.43) & \tiny(4.14) & \tiny(4.08) & \tiny(1.33) & \tiny(2.76) & \textbf{\tiny(2.58)} & \tiny(2.84) \\
 & 21.40 & 26.03 & 20.37 & 10.68 & 12.36 & 11.58 & \textbf{5.68} & 6.99 & 15.30 \\
\multirow{-2}{*}{I} & \tiny(2.69) & \tiny(3.01) & \tiny(1.50) & \tiny(4.14) & \tiny(4.49) & \tiny(2.12) & \textbf{\tiny(2.43)} & \tiny(3.03) & \tiny(3.00) \\
\rowcolor{gray!25}
 & 10.02 & 10.60 & 14.64 & 15.96 & 22.23 & 12.13 & 7.32 & \textbf{4.25} & 8.22 \\
\rowcolor{gray!25}
\multirow{-2}{*}{A} & \tiny(5.34) & \tiny(5.66) & \tiny(4.02) & \tiny(8.28) & \tiny(5.14) & \tiny(3.82) & \tiny(2.50) & \textbf{\tiny(3.13)} & \tiny(3.23) \\
 & 15.78 & 12.28 & 15.73 & 12.40 & 16.61 & 14.04 & \textbf{7.14} & 16.46 & 20.59 \\
\multirow{-2}{*}{B} & \tiny(3.93) & \tiny(3.67) & \tiny(2.41) & \tiny(7.05) & \tiny(4.68) & \tiny(2.62) & \textbf{\tiny(2.78)} & \tiny(3.99) & \tiny(2.28) \\
\rowcolor{gray!25}
 & 14.86 & 12.64 & 11.79 & 12.02 & 15.25 & 11.32 & 4.26 & \textbf{2.03} & 10.74 \\
\rowcolor{gray!25}
\multirow{-2}{*}{C} & \tiny(4.82) & \tiny(4.63) & \tiny(2.09) & \tiny(5.71) & \tiny(3.92) & \tiny(2.07) & \tiny(2.30) & \textbf{\tiny(1.63)} & \tiny(3.63) \\
 & 7.93 & 49.57 & 11.87 & \multirow{2}{*}{--} & \multirow{2}{*}{--} & 11.12 & 5.49 & \textbf{3.34} & 6.48 \\
\multirow{-2}{*}{H} & \tiny(3.28) & \tiny(1.05) & \tiny(2.46) &  &  & \tiny(1.97) & \tiny(2.34) & \textbf{\tiny(3.76)} & \tiny(3.76) \\
\rowcolor{gray!25}
 & 8.32 & 10.88 & 12.41 & 13.28 & 18.96 & 7.42 & 4.11 & 3.19 & \textbf{2.85} \\
\rowcolor{gray!25}
\multirow{-2}{*}{L} & \tiny(5.72) & \tiny(5.79) & \tiny(4.58) & \tiny(9.12) & \tiny(5.85) & \tiny(3.34) & \tiny(2.54) & \tiny(2.75) & \textbf{\tiny(1.40)} \\
 & 50.02 & 12.20 & 10.74 & 13.46 & 18.01 & 8.47 & 4.98 & 2.65 & \textbf{2.56} \\
\multirow{-2}{*}{P} & \tiny(8.59) & \tiny(5.76) & \tiny(3.88) & \tiny(5.32) & \tiny(4.34) & \tiny(2.81) & \tiny(2.48) & \tiny(2.08) & \textbf{\tiny(0.96)} \\
\rowcolor{gray!25}
 & 24.19 & 14.46 & 15.44 & 6.46 & 10.47 & 7.86 & 4.17 & \textbf{2.64} & 11.49 \\
\rowcolor{gray!25}
\multirow{-2}{*}{R} & \tiny(11.21) & \tiny(4.73) & \tiny(2.94) & \tiny(4.77) & \tiny(3.07) & \tiny(2.13) & \tiny(2.57) & \textbf{\tiny(2.09)} & \tiny(3.29) \\
 & 12.12 & 21.98 & 16.20 & 5.45 & 8.77 & 8.56 & 3.23 & \textbf{2.23} & 4.95 \\
\multirow{-2}{*}{W} & \tiny(5.31) & \tiny(4.86) & \tiny(2.37) & \tiny(3.91) & \tiny(4.28) & \tiny(1.97) & \tiny(2.46) & \textbf{\tiny(1.87)} & \tiny(2.21) \\
\hline
$\overline{\textrm{rank}}$ & 6.9 & 6.3 & 6.6 & 5.7 & 7.5 & 4.4 & 2.0 & 1.9 & 3.8 \\
\hline                    
\end{tabular}
\caption{Mean absolute error (standard deviation in parentheses), in percentages, for experiment \#3-b (75-percentile of performance among negative sub-classes). Results for datasets C, R and W are repetitions from Table~\ref{tab:err:xp:basic}, since they contain only one negative sub-class.}
\label{tab:err:xp:75}
\end{table}


\begin{table}[htb]
\centering
\scriptsize
\begin{tabular}{r|cccccccc|c}
\hline
Data & EN\textsubscript{a} & EN\textsubscript{s} & PE & KM1 & KM2 & TIcE & ExTIcE & PAT\textsubscript{M} & BFT\textsubscript{M} \\
\hline
\rowcolor{gray!25}
 & 49.87 & 18.89 & 38.49 & 18.07 & 22.24 & 17.18 & \textbf{7.69} & 49.95 & 44.11 \\
\rowcolor{gray!25}
\multirow{-2}{*}{N} & \tiny(10.17) & \tiny(6.36) & \tiny(1.07) & \tiny(4.09) & \tiny(4.99) & \tiny(3.17) & \textbf{\tiny(3.24)} & \tiny(0.76) & \tiny(1.09) \\
 & 28.14 & 38.08 & 24.81 & 16.41 & 18.50 & 22.16 & \textbf{12.67} & 38.28 & 33.04 \\
\multirow{-2}{*}{I} & \tiny(3.27) & \tiny(3.97) & \tiny(2.41) & \tiny(6.12) & \tiny(6.22) & \tiny(3.02) & \textbf{\tiny(3.22)} & \tiny(4.42) & \tiny(3.24) \\
\rowcolor{gray!25}
 & 10.83 & 11.23 & 15.18 & 19.91 & 25.78 & 15.03 & 9.79 & \textbf{5.68} & 12.92 \\
\rowcolor{gray!25}
\multirow{-2}{*}{A} & \tiny(5.65) & \tiny(6.57) & \tiny(6.71) & \tiny(9.95) & \tiny(5.35) & \tiny(4.01) & \tiny(2.95) & \textbf{\tiny(3.67)} & \tiny(3.73) \\
 & 19.99 & 17.19 & 18.57 & 18.90 & 21.33 & 17.48 & \textbf{9.10} & 16.59 & 21.98 \\
\multirow{-2}{*}{B} & \tiny(3.92) & \tiny(3.91) & \tiny(2.25) & \tiny(8.21) & \tiny(5.93) & \tiny(2.76) & \textbf{\tiny(2.81)} & \tiny(3.57) & \tiny(2.87) \\
\rowcolor{gray!25}
 & 14.86 & 12.64 & 11.79 & 12.02 & 15.25 & 11.32 & 4.26 & \textbf{2.03} & 10.74 \\
\rowcolor{gray!25}
\multirow{-2}{*}{C} & \tiny(4.82) & \tiny(4.63) & \tiny(2.09) & \tiny(5.71) & \tiny(3.92) & \tiny(2.07) & \tiny(2.30) & \textbf{\tiny(1.63)} & \tiny(3.63) \\
 & 7.93 & 49.57 & 11.87 & \multirow{2}{*}{--} & \multirow{2}{*}{--} & 11.12 & 5.49 & \textbf{3.34} & 6.48 \\
\multirow{-2}{*}{H} & \tiny(3.28) & \tiny(1.05) & \tiny(2.46) &  &  & \tiny(1.97) & \tiny(2.34) & \textbf{\tiny(3.76)} & \tiny(3.76) \\
\rowcolor{gray!25}
 & 8.81 & 12.95 & 13.11 & 16.86 & 23.82 & 9.10 & 5.83 & \textbf{3.65} & 19.83 \\
\rowcolor{gray!25}
\multirow{-2}{*}{L} & \tiny(6.07) & \tiny(5.69) & \tiny(5.09) & \tiny(9.60) & \tiny(5.79) & \tiny(3.59) & \tiny(3.16) & \textbf{\tiny(2.47)} & \tiny(6.07) \\
 & 50.40 & 13.71 & 11.14 & 14.69 & 19.43 & 11.74 & 6.65 & \textbf{2.85} & 13.29 \\
\multirow{-2}{*}{P} & \tiny(1.81) & \tiny(5.99) & \tiny(3.66) & \tiny(6.29) & \tiny(5.83) & \tiny(3.75) & \tiny(3.02) & \textbf{\tiny(2.25)} & \tiny(6.74) \\
\rowcolor{gray!25}
 & 24.19 & 14.46 & 15.44 & 6.46 & 10.47 & 7.86 & 4.17 & \textbf{2.64} & 11.49 \\
\rowcolor{gray!25}
\multirow{-2}{*}{R} & \tiny(11.21) & \tiny(4.73) & \tiny(2.94) & \tiny(4.77) & \tiny(3.07) & \tiny(2.13) & \tiny(2.57) & \textbf{\tiny(2.09)} & \tiny(3.29) \\
 & 12.12 & 21.98 & 16.20 & 5.45 & 8.77 & 8.56 & 3.23 & \textbf{2.23} & 4.95 \\
\multirow{-2}{*}{W} & \tiny(5.31) & \tiny(4.86) & \tiny(2.37) & \tiny(3.91) & \tiny(4.28) & \tiny(1.97) & \tiny(2.46) & \textbf{\tiny(1.87)} & \tiny(2.21) \\
\hline
$\overline{\textrm{rank}}$ & 6.4 & 6.0 & 5.9 & 5.5 & 7.0 & 4.2 & 1.7 & 2.7 & 5.6 \\
\hline                    
\end{tabular}
\caption{Mean absolute error (standard deviation in parentheses), in percentages, for Experiment \#3-c (worst performance among negative sub-classes). Results for datasets C, R and W are repetitions from Table~\ref{tab:err:xp:basic}, since they contain only one negative sub-class.}
\label{tab:err:xp:worst}
\end{table}


\begin{figure}[htb]
\centering
\subfloat[Experiment \#3-b.\label{fig:nemenyi:75}]{\includegraphics[width=.45\columnwidth]{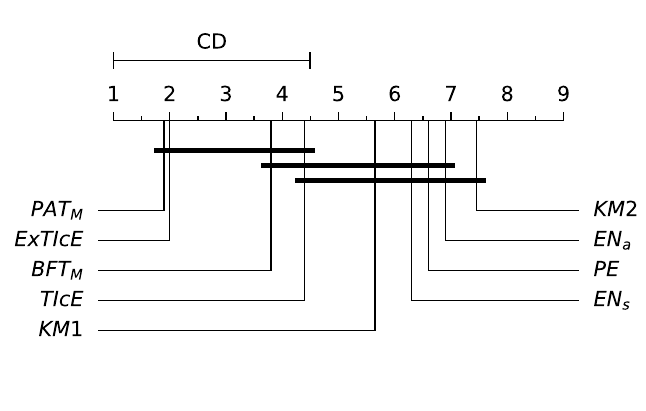}}
\hspace{0.3cm}
\subfloat[Experiment \#3-c.\label{fig:nemenyi:worst}]{\includegraphics[width=.45\columnwidth]{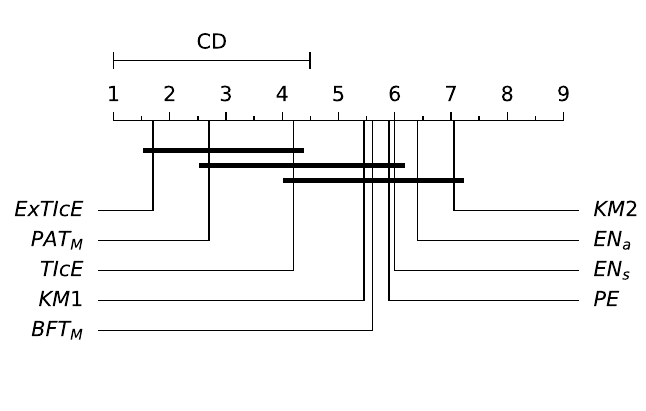}}
\vspace{0.3cm}
\caption{Nemenyi test for Experiments \#3-b and \#3-c with $\alpha=0.1$. Methods within the \textbf{c}ritical \textbf{d}ifference (CD) are connected by horizontal line and are not significantly different.}
\label{fig:nemenyi:75w}
\end{figure}

Table~\ref{tab:err:xp:75} and Figure~\ref{fig:nemenyi:75} present the results for Experiment \#3-b, and Table~\ref{tab:err:xp:worst} and Figure~\ref{fig:nemenyi:worst}, for Experiment \#3-c. Whereas for  75-percentile (Experiment \#3-b) PAT\textsubscript{M} still maintains significantly lower MAE than ExTIcE in pairwise comparison (p-value of $0.03$ according to Wilcoxon Rank-Sum test), the opposite takes place for  100-percentile (Experiment \#3-c). In fact, due to the poor performance of PAT\textsubscript{M} in datasets N, I, and B, its average rank was $2.7$ despite the fact that the algorithm ranked first in all other datasets. Finally, although the performance of ExTIcE for the same datasets decreased in comparison to the previous experiments, it still outperformed all other approaches. In the remaining datasets, ExTIcE ranked second, only behind PAT\textsubscript{M}.

Particularly for dataset N, observe that the average error obtained by PAT\textsubscript{M} is close to 50\% in Table~\ref{tab:err:xp:worst}. As the actual positive ratio varied uniformly within the interval $[0,1]$ during the experiment, such an error indicates that PAT\textsubscript{M} always predicted $\hat{p}$ as either close to zero or close to one. Considering our previous results for PAT\textsubscript{M} in this same dataset, we can infer that the current situation corresponds to the latter case, since the algorithm could previously detect situations where the positive class was not prominent (the error was below the baseline 25\%), and the learning process involved only the positive class. In fact, further analysis of our more detailed data (available as supplemental material \citep{supmatweb}) reveals that the average prediction of $p$ was $99.74\%$, which indicates that observations from the negative class obtained score values at least as large as those of positive observations. From this piece of data, we can assume that observations that belong to this negative class are highly similar to at least part of the positive data, fact that also affected the best classify and count  BFT\textsubscript{M}. In the next section, we discuss how and why this scenario affected PAT\textsubscript{M} to a considerably greater degree than ExTIcE. Before that, we do our final analysis regarding time consumption.


\begin{table}[htb]
\rowcolors{2}{}{gray!25}
\centering
\scriptsize
\begin{tabular}{r|ccccccc|c}
\hline
Data & EN\textsubscript{a} & EN\textsubscript{s} & PE & KM & TIcE & ExTIcE & PAT\textsubscript{M} & PAT\textsubscript{M} w/o T \\
\hline
N & 20.19 & 8.41 & 2960.44 & 227.17 & 4.50 & 84.59 & \textbf{2.55} & 1.82 \\
I & 15.80 & 6.67 & 2971.05 & 240.50 & 13.68 & 232.96 & \textbf{2.58} & 1.84 \\
A & 1.30 & 1.27 & 212.41 & 23.12 & 4.78 & 142.62 & \textbf{1.03} & 0.67 \\
B & 6.53 & 9.89 & 2959.61 & 209.72 & 6.15 & 110.52 & \textbf{2.61} & 1.79 \\
C & 5.30 & 6.44 & 1432.50 & 213.06 & 7.89 & 139.28 & \textbf{2.16} & 1.65 \\
H & 9.86 & 5.56 & 1231.82 & -- & 23.78 & 339.50 & \textbf{2.17} & 1.47 \\
L & 1.08 & \textbf{0.78} & 253.69 & 15.10 & 3.72 & 95.65 & 1.05 & 0.68 \\
P & 4.07 & 1.40 & 518.00 & 41.02 & 4.48 & 114.10 & \textbf{1.37} & 0.94 \\
R & 8.26 & 2.53 & 773.14 & 150.89 & 2.35 & 45.23 & \textbf{1.68} & 1.16 \\
W & 6.01 & 2.07 & 745.12 & 125.60 & 3.17 & 59.29 & \textbf{1.58} & 1.22 \\
\hline
\rowcolor{white}
$\overline{\textrm{rank}}$ & 4.3 & 3.4 & 8.0 & 6.6 & 4.2 & 6.4 & 2.1 & 1.0 \\
\hline                 
\end{tabular}
\caption{Total time spent, in seconds, to accomplish all tasks in  experiment \#4. As PAT\textsubscript{M} is the only algorithm that produces a model that can be reused for several test samples, one additional column shows the time spent by PAT\textsubscript{M} disregarding the training stage.}
\label{tab:time:xp}
\end{table}


Table~\ref{tab:time:xp} presents the total time, in seconds, required to perform all tasks in Experiment \#4. We can see that PE was several orders of magnitude slower than the other approaches. KM and predictably ExTIcE were  both orders of magnitude slower than TIcE, PAT\textsubscript{M} and EN. Although TIcE, EN and PAT\textsubscript{M} were generally in the same order of magnitude, PAT\textsubscript{M} performed consistently faster, even when the time necessary to train the scorer is considered.

Given the proposed experimental setup, we cannot conclusively claim that EN, TIcE and PAT always have  numerically similar execution times. We note that the training dataset was limited to 500 observations, and the test sample to 2,000. We believe that further experimentation would have shown both EN and TIcE to become several orders of magnitude slower than PAT\textsubscript{M} for bigger samples due to the time complexities of SVM and TIcE. Additionally, replacing PAT\textsubscript{M}'s Mahalanobis Distance with a different dissimilarity function would also impact its execution time performance.




%% file: sections/discussion.tex
\section{Discussion}
\label{sec:discussion}

Elkan's method (EN) has historical value as it puts the spotlight on Positive and Unlabeled Prior Estimation, a problem that is similar to One-class Quantification. EN also introduced theoretical basis for newer algorithms to improve on. However, as the results of our data-driven experimentation showed, such a method usually presented a poor performance. 

In the previous context, we would not to recommend EN as a first-choice method to address a quantification task. Nevertheless, given that EN is a classical method that can achieve one-class quantification, we argue that it should be used as a baseline when comparing other methods. We would neither recommend PE, since our experiments demonstrated there is no statistical evidence of the difference of performance between PE and EN. In addition to that, PE was shown to be the slowest approach among all algorithms tested. 

As explained in Section~\ref{sec:exp:algs}, BFT represents the best possible Classify and Count derived from a one-class scorer. However, as discussed in Section~\ref{sec:def:quant}, due to the systematic error of CC, in practice, BFT will tend to be outperformed by the other methods. Despite this, like EN, we argue that this method can be used as baseline in the comparison of novel quantification algorithms. 

ExTIcE fulfilled its role of showing the potential of TIcE's underlying search problem. Indeed, the former consistently provided smaller absolute quantification errors than the latter. Nevertheless, our purpose is not to defend ExTIcE's position and recommend it as a quantifier, but rather entice the community to further explore the region search problem proposed by TIcE, in future work. ExTIcE, while less restricted than TIcE, is still limited in a number of ways. For instance, like in TIcE and most other tree algorithms, the sub-regions explored only \textit{``cut''} the feature space along its axes. Additionally, we believe it is possible to create an algorithm from the ideas of TIcE that, similarly to PAT, is capable of inducing a model that can later be used to quantify several test samples without resorting to the training data.

\cite{ramaswamy2016mixture} make the argument that ``requiring
an accurate conditional probability estimate (which is a real
valued function over the feature space) for estimating the
mixture proportion (a single number) is too roundabout''. On the other hand, we defend that the referred approach is actually very practical, since there are already a number of methods for this exactly purpose that are accessible for even inexperienced practitioners. This approach is also the base of PAT, which is, in our opinion, notoriously simpler than KM, yet generally providing smaller quantification errors at an unquestionable faster rate.

In our experiments, PAT was shown to produce the smallest quantification errors while being the fastest algorithm. For this reason, it is the algorithm we mostly recommend for practical use.

Notwithstanding the favorable results, we must highlight PAT's drawbacks, which were evidenced by the evolution of Experiment \#3. PAT was developed on the assumption that some negative observations can be similar to positive observations up to a certain degree. The algorithm (indirectly) tries to ignore the presence of negative observations close to the boundaries of the positive class, in the feature space, by extrapolating the number of observations from only the top scored ones.

However, consider the case where a negative sub-class is partially identical to the positive class, in the sense that a \textit{number of} negative observations are, \textit{individually}, \textit{identical to} or \textit{indistinguishable from} positive observations. In such a case, the quantification of PAT will likely be affected, since PAT does its computations solely on the observations' scores. Naturally, the degree to which PAT will be affected depends on the proportion of the aforementioned sub-class within the negative class.

\begin{figure}[htb]
\centering
\subfloat[\textit{Training data}.\label{fig:tice:bt:pat:tr}]{\includegraphics[width=.45\columnwidth]{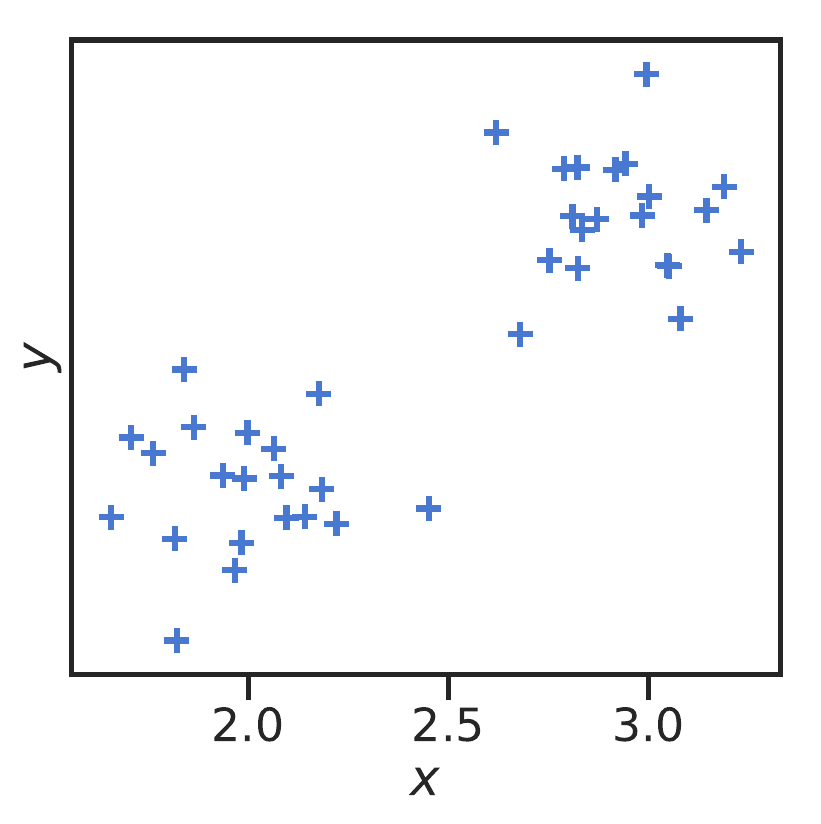}}
\hspace{0.3cm}
\subfloat[\textit{Test data}.\label{fig:tice:bt:pat:test}]{\includegraphics[width=.45\columnwidth]{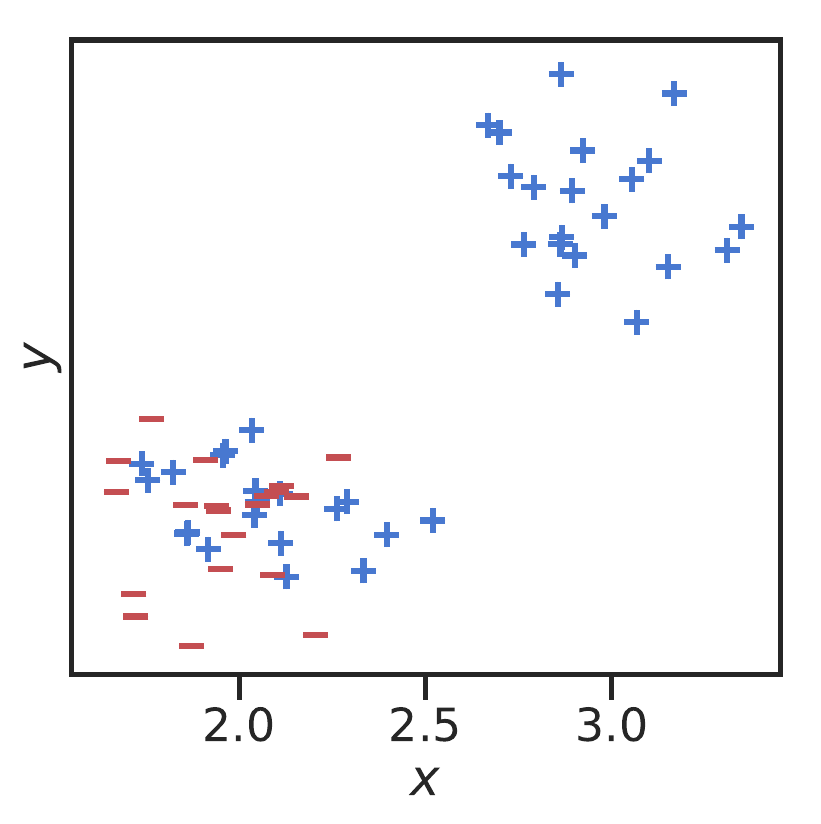}}
\vspace{0.3cm}
\caption{A hypothetical situation where the main approach of TIcE would lead to better quantification than PAT's approach, regardless of the underlying scorer of the latter. Blue $+$ (plus) symbol indicates positive observations, while red $-$ (minus) symbol indicates negative observations.}
\label{fig:tice:bt:pat}
\end{figure}

Meanwhile, ExTIcE could be less or not affected by those partially identical classes. Indeed, its search mechanism allows it to completely ignore regions of the feature space where such overlaps are more prevalent, \textit{if} there are other regions with less overlap. Figure~\ref{fig:tice:bt:pat} illustrates this discussion. Notice that ExTIcE would likely only consider the top-right quadrant of the feature space to infer $\mathfrak{c}$, while PAT would use all scores, even though negative observations are as highly scored as positive observations, in this scenario.

Remarks being done for PAT and ExTiCE, we argue that, for practical reasons, the overlaps mentioned may indicate a need to revise: \textbf{(a)} whether the negative observations actually \textit{should} or \textit{need} be classified as negative and \textbf{(b)} the quality of the existing features.

In any case, we can try to minimize the effects of negative classes that are identical to positive observations on PAT by ensembling it along with ExTIcE. In this particular scenario, PAT overestimates $\hat{p}$ as a result of negative observations being considered to be positive. In addition to that, we noticed that ExTIcE tends to generally overestimate $\hat{p}$. The latter finding is not straightforward: since ExTIcE biasedly tries to maximize a \textit{lower bound} of $\hat{c}_\gamma$ (the proportion of labeled data over all positive data). It may seem that we would actually expect $\hat{p}$ (the proportion of unlabeled positive data over all unlabeled data) to be underestimated. However, ExTIcE overestimated 75\% of its predictions in Experiment \#3 (considering all classes). Such an overestimation can be justified by correction factor, which makes the method search for \textit{local lower bounds}, being too heavy. The occasional heavy overestimation of PAT along with the general overestimation of ExTIcE favor the approach of considering the minimum between the predictions provided by both methods. 

\begin{table}[htb]
\centering
\scriptsize
\begin{tabular}{r|ccc|ccc}
\hline
& \multicolumn{3}{c|}{Experiment \#3-a} & \multicolumn{3}{c}{Experiment \#3-c} \\
Data & ExTIcE & PAT\textsubscript{M} & $\min\{\hat{p}\}$ & ExTIcE & PAT\textsubscript{M} & $\min\{\hat{p}\}$ \\
\hline
\rowcolor{gray!25}
 & 3.53 & \textbf{2.09} & 2.31 & 7.69 & 49.95 & \textbf{7.59} \\
\rowcolor{gray!25}
\multirow{-2}{*}{N} & \tiny(1.81) & \textbf{\tiny(1.75)} & \tiny(1.69) & \tiny(3.24) & \tiny(0.76) & \textbf{\tiny(3.19)} \\
 & 5.23 & \textbf{2.16} & 2.17 & 12.67 & 38.28 & \textbf{12.60} \\
\multirow{-2}{*}{I} & \tiny(2.35) & \textbf{\tiny(1.67)} & \tiny(1.70) & \tiny(3.22) & \tiny(4.42) & \textbf{\tiny(2.94)} \\
\rowcolor{gray!25}
 & 6.44 & \textbf{3.73} & 4.20 & 9.79 & 5.68 & \textbf{5.59} \\
\rowcolor{gray!25}
\multirow{-2}{*}{A} & \tiny(2.28) & \textbf{\tiny(2.69)} & \tiny(2.32) & \tiny(2.95) & \tiny(3.67) & \textbf{\tiny(2.93)} \\
 & 4.89 & 3.21 & \textbf{2.86} & 9.10 & 16.59 & \textbf{8.12} \\
\multirow{-2}{*}{B} & \tiny(2.17) & \tiny(1.99) & \textbf{\tiny(1.76)} & \tiny(2.81) & \tiny(3.57) & \textbf{\tiny(2.45)} \\
\rowcolor{gray!25}
 & 4.26 & \textbf{2.03} & 2.12 & 4.26 & \textbf{2.03} & 2.12 \\
\rowcolor{gray!25}
\multirow{-2}{*}{C} & \tiny(2.30) & \textbf{\tiny(1.63)} & \tiny(1.70) & \tiny(2.30) & \textbf{\tiny(1.63)} & \tiny(1.70) \\
 & 5.49 & \textbf{3.34} & 4.44 & 5.49 & \textbf{3.34} & 4.44 \\
\multirow{-2}{*}{H} & \tiny(2.34) & \textbf{\tiny(3.76)} & \tiny(6.96) & \tiny(2.34) & \textbf{\tiny(3.76)} & \tiny(6.96) \\
\rowcolor{gray!25}
 & 3.73 & \textbf{3.07} & 3.26 & 5.83 & 3.65 & \textbf{3.54} \\
\rowcolor{gray!25}
\multirow{-2}{*}{L} & \tiny(2.53) & \textbf{\tiny(2.38)} & \tiny(2.56) & \tiny(3.16) & \tiny(2.47) & \textbf{\tiny(2.61)} \\
 & 4.62 & \textbf{2.58} & 3.43 & 6.65 & \textbf{2.85} & 3.59 \\
\multirow{-2}{*}{P} & \tiny(2.68) & \textbf{\tiny(2.11)} & \tiny(2.52) & \tiny(3.02) & \textbf{\tiny(2.25)} & \tiny(2.41) \\
\rowcolor{gray!25}
 & 4.17 & \textbf{2.64} & 2.81 & 4.17 & \textbf{2.64} & 2.81 \\
\rowcolor{gray!25}
\multirow{-2}{*}{R} & \tiny(2.57) & \textbf{\tiny(2.09)} & \tiny(2.17) & \tiny(2.57) & \textbf{\tiny(2.09)} & \tiny(2.17) \\
 & 3.23 & \textbf{2.23} & 2.67 & 3.23 & \textbf{2.23} & 2.67 \\
\multirow{-2}{*}{W} & \tiny(2.46) & \textbf{\tiny(1.87)} & \tiny(1.93) & \tiny(2.46) & \textbf{\tiny(1.87)} & \tiny(1.93) \\
\hline
$\overline{\textrm{rank}}$ & 3.0 & 1.1 & 1.9 & 2.7 & 1.8 & 1.5 \\
\hline                    
\end{tabular}
\caption{Mean absolute error (standard deviation in parentheses) of ExTIcE, PAT\textsubscript{M}, and an ensemble with both methods. Results are presented in percentages and include for Experiments \#3--a (median of performance among negative sub-classes) and \#3--c (worst performance among negative sub-classes).}
\label{tab:err:xp:tpm}
\end{table}

Table~\ref{tab:err:xp:tpm} presents the results of experiments \#3--a and \#3--c for ExTIcE, PAT\textsubscript{M} and an ensemble that outputs the minimum prediction between the two methods. In experiment \#3--a (median), we can see that, for all but one dataset, the ensemble performs better than ExTIcE and worse than PAT\textsubscript{M}. In the exceptional case of dataset A, the ensemble performed better than the other methods. We note that in all but one dataset, the performance of the ensemble was numerically closer to the performance of PAT\textsubscript{M} rather than ExTIcE. On the other hand, in experiment \#3--c, the ensemble has a better performance than PAT\textsubscript{M} in multiple datasets. Differently from PAT\textsubscript{M}, the ensemble could perform well in the problematic datasets N, I and B. However, we emphasize that this ensemble imposes a high computational cost due to the use of ExTIcE. Our main purpose is to highlight that it is indeed possible to achieve performance similar to PAT's while handling the particular case where it cannot perform well. We expect other, faster, methods to be developed in future work.

Finally, both PAT and ExTIcE strongly depend on the assumption that the distribution of the positive class is the same in both training and test samples. Given their strategies, we safely presume that they would be severely affected in the event of the assumption being false. 

%% file: sections/conclusion.tex
\section{Conclusion}
\label{sec:conclusion}

In this paper, we described several distinct approaches for the one-class quantification problem, most of which are derived from the area of research known as Positive and Unlabeled Learning.

We empirically showed the superiority of our proposal, Passive Aggressive Threshold (PAT) for one-class quantification problems, given that the distribution of the negative class is unknown and overlap with the positive class is allowed only up to a reasonable degree. However, we stress that PAT performs poorly in cases where a reasonable portion of the negative class is indistinguishable from positive observation points. 

We also showed how the region search optimization problem behind Tree Induction for $\mathfrak{c}$ Estimation (TIcE) is able to solve one-class quantification tasks in which a portion of the negative observations can be identical to or indistinguishable from positive observations. However, such an approach still requires further development, as we demonstrated with our superior (in terms of lower quantification error) version ExTIcE.

For future work, we are interested in exploring better one-class scorers for PAT, and develop methods to solve the search problem proposed by TIcE. Additionally, on the latter objective, we aim to develop methods that can train solely with positive observations and later quantify several independent test samples, to qualify as One-class Quantification algorithms.

%% file: sections/acknowledgements.tex
\section*{Acknowledgements}
This study was financed in part by the Coordena\c{c}\~ao de Aperfei\c{c}oamento de Pessoal de N\'ivel Superior - Brasil (CAPES) - Finance Code PROEX-6909543/D, the Funda\c{c}\~ao de Amparo a Pesquisa do Estado de S\~ao Paulo (FAPESP, grant 2017/22896-7), and the United States Agency for International Development (USAID, grant AID-OAA-F-16-00072).

%% file: sections/appendix.tex
\begin{appendices}
\section{Analysis on TIcE's time complexity}
\label{appen:tices:complexity}
In this section, we thoroughly analyze the time complexity of TIcE. For this analysis, consider only binary nominal attributes and splits at the median for numerical attributes, so that the data is always split into two slices.

To evaluate the goodness for each attribute, when splitting a node $i$, it is necessary to count how many positive observations go to each side after the split. To this end, the code provided by \cite{bekker2018estimating} uses a data structure called \textit{BitArray} for such a counting. The structure is instantiated and initialized for each possible split. Although \textit{BitArray} is highly optimized, especially so for memory usage, it still performs the counting in $O(n_i)$, where $n_i$ is the number of observations assessed by the splitting node. Additionally, any data structure that is below $O(n_i)$ for exact counting would still have an initialization that is $\Omega(n_i)$, since every observation must be processed to give enough information to the structure about the counting. $O(n_i)$ is the same time complexity of a linear counting using a standard array. The authors do not comment on alternatives.

We note that it is possible to use another data structure, like binary decision trees, to obtain the count in $O(\log n_i)$. This data structure can be updated after the split for the attribute that caused the split: for numerical attributes, this can be done with a Cartesian tree in $O(\log n_i)$, and for binary nominal attributes, this is not necessary since the attribute should not be used any longer. However, for the remaining attributes, there is no such a way to quickly place each observation into the correct side of the split, since no relation between the splitting attribute and the other attributes is guaranteed. Therefore, the data structure for each attribute should be updated, resulting in a $O(m_in_i\log n_i)$ for the split when using such a data structure, where $m_i$ is the number of attributes that the node has access to. On the other hand, by ditching this data structure, the split is a lower $O(m_in_i)$, { so this is the complexity is what we consider from now on}.

{Assuming} that each attribute is used only once,{ that} the data is always split in half { (optimistic scenario), and that} \textit{there is enough data}, the maximum height is $m$, \emph{i.e.}, the total number of attributes, and the complexity of the algorithm is $O(2^mn)$, as shown in Equation~\ref{eq:tice:complexity}, where $n$ is the total number of observations, and $F(m,n)$ is the recurrence relation of the algorithm.
    
    \begin{align}
        \label{eq:tice:complexity}
    \begin{split}
        F(h, n) &= mn + 2F\left(m - 1, \frac{n}{2}\right) \\
        &=mn + 2\left(
         \left(m - 1\right)\frac{n}{2} + 2F\left(m - 2, \frac{n}{4}\right)
        \right)\\
        &= mn + 2\left(m - 1\right)n + 4F\left(m - 2,\frac{n}{4}\right)\\
        &\ldots\\
        &= mn + 2\left(m - 1\right)n + \ldots + 2^{m - 1}\left(m - m + 1\right)n\\
        &= n\sum_{k=0}^{m}{2^k\left(m - k\right)}\\
        &= 2n(-m + 2^m - 1)\\
        &O\left(2^mn\right)
    \end{split}
    \end{align}
    
If there is not enough data to use all attributes, but, again, each attribute is used \textit{only once} and data {is always halved}, the complexity is $O(n^2)$, since the maximum height is $\log_2n$ and $2^{\log_2n} = n$. Therefore, the general complexity of the algorithm is $O(\min(n^2, 2^mn))$ when each feature is used only once and the data is divided in half, which is significantly higher than the $O(mn)$ stated by \cite{bekker2018estimating}. If the attributes can be used more than once and/or the data is not evenly divided after each split, the complexity is even higher. This fact emphasizes the overly optimistic initial assessment of \cite{bekker2018estimating}.

\section{Additional experimental results}
\label{appen:add:exp:results}

\subsection{Random Forest for c Estimation}
\label{appen:ranfoce}

{ In Section~\ref{sec:tice}, we noticed that the search of TIcE is biased and briefly explained the mechanisms by which the method prevents an overestimation of $\mathfrak{c}$. Here, we present an alternative, yet similar, search method that deal with the biases differently and more naively.}

To this end, we developed a baseline method that we call Random Forest for $\mathfrak{c}$ Estimation (RanFocE). Consider the minimum node size $l$. Each tree $T$ has its nodes split randomly: for each node, one feature that can split its corresponding data into two sets with \textit{at least} $l$ \textit{labeled} data points is chosen, and the data is split according to said feature at a random threshold value under the same constraint. If no such a feature exists, the node is split no more. The estimation of $\mathfrak{c}$ for a node $n$ is $\hat{\mathfrak{c}}_n$ and is the total number of labeled data points divided by the total number of data points. The estimation of $\mathfrak{c}$ for a tree is $\hat{\mathfrak{c}}_T = \max_{n \in T}\hat{\mathfrak{c}}_n$. The final estimation is the median of the estimates provided by all trees in the forest. We note that, contrary to TIcE, no correction is applied to the estimations to transform them into lower bounds.

In our experiments, we induced $100$ trees per forest. We adopted a value for $l$ similar to the minimum number of data points in TIcE: $l = \min\{1000, \lfloor 0.5 + 0.1\times |L| \rfloor\}$.

Our empirical results under the settings proposed by Experiment \#3-a and Experiment \#3-c, which are described in Section~\ref{sec:exp:setup}, are shown in Table~\ref{tab:ranfoce}. With this data, we cannot infer statistical difference between the algorithms in either setting ($p$-values of $0.36$ and $0.41$ for experiments \#3-a and \#3-c, respectively, according to a Wilcoxon Rank-Test). Therefore, we cannot conclude that TIcE performs better than a similar tree induction algorithm that splits randomnly and do not perform a correction to the final estimation of $\mathfrak{c}$.

\begin{table}[htb]
    \centering
\caption[Experimental RanFoCE]{Mean absolute error (standard deviation in parentheses) of TIcE and RanFocE. Results are presented in percentages and include for experiments \#3--a and \#3--c.}
\scriptsize
\begin{tabular}{r|cc|cc}
\hline
& \multicolumn{2}{c|}{Experiment \#3-a} & \multicolumn{2}{c}{Experiment \#3-c} \\
Data & TIcE & RanFocE & TIcE & RanFocE \\
\hline
\rowcolor{gray!25}
 & \textbf{7.93} & 8.25 & 17.18 & \textbf{14.62} \\
\rowcolor{gray!25}
\multirow{-2}{*}{N} & \textbf{\tiny(1.27)} & \tiny(1.15) & \tiny(3.17) & \textbf{\tiny(2.09)} \\
 & 10.92 & \textbf{5.96} & 22.16 & \textbf{9.76} \\
\multirow{-2}{*}{I} & \tiny(1.42) & \textbf{\tiny(1.38)} & \tiny(3.02) & \textbf{\tiny(2.98)} \\
\rowcolor{gray!25}
 & 10.81 & \textbf{10.14} & 15.03 & \textbf{10.97} \\
\rowcolor{gray!25}
\multirow{-2}{*}{A} & \tiny(3.59) & \textbf{\tiny(1.99)} & \tiny(4.01) & \textbf{\tiny(1.64)} \\
 & 11.22 & \textbf{6.04} & 17.48 & \textbf{9.09} \\
\multirow{-2}{*}{B} & \tiny(2.12) & \textbf{\tiny(1.11)} & \tiny(2.76) & \textbf{\tiny(1.81)} \\
\rowcolor{gray!25}
 & 11.32 & \textbf{6.18} & 11.32 & \textbf{6.18} \\
\rowcolor{gray!25}
\multirow{-2}{*}{C} & \tiny(2.07) & \textbf{\tiny(1.31)} & \tiny(2.07) & \textbf{\tiny(1.31)} \\
 & 11.12 & \textbf{7.16} & 11.12 & \textbf{7.16} \\
\multirow{-2}{*}{H} & \tiny(1.97) & \textbf{\tiny(1.27)} & \tiny(1.97) & \textbf{\tiny(1.27)} \\
\rowcolor{gray!25}
 & \textbf{6.53} & 12.63 & \textbf{9.10} & 13.69 \\
\rowcolor{gray!25}
\multirow{-2}{*}{L} & \textbf{\tiny(2.86)} & \tiny(2.24) & \textbf{\tiny(3.59)} & \tiny(2.28) \\
 & \textbf{7.84} & 11.25 & \textbf{11.74} & 12.20 \\
\multirow{-2}{*}{P} & \textbf{\tiny(3.18)} & \tiny(1.99) & \textbf{\tiny(3.75)} & \tiny(2.12) \\
\rowcolor{gray!25}
 & \textbf{7.86} & 9.54 & \textbf{7.86} & 9.54 \\
\rowcolor{gray!25}
\multirow{-2}{*}{R} & \textbf{\tiny(2.13)} & \tiny(2.12) & \textbf{\tiny(2.13)} & \tiny(2.12) \\
 & 8.56 & \textbf{8.23} & 8.56 & \textbf{8.23} \\
\multirow{-2}{*}{W} & \tiny(1.97) & \textbf{\tiny(1.42)} & \tiny(1.97) & \textbf{\tiny(1.42)} \\
\hline
$\overline{\textrm{rank}}$ & 1.6 & 1.4 & 1.7 & 1.3 \\
\hline
    \end{tabular}
    \label{tab:ranfoce}
\end{table}

\subsection{Exp \#2}
\label{appen:exp:2}

In Table~\ref{tab:err:xp:rnd} we present the results obtained in Experiment \#2, which were omitted in Section~\ref{sec:exp:eval}.

\begin{table}[htbp]
\centering
\scriptsize
\begin{tabular}{r|cccccccc|c}
\hline
Data & EN\textsubscript{a} & EN\textsubscript{s} & PE & KM1 & KM2 & TIcE & ExTIcE & PAT\textsubscript{M} & BFT\textsubscript{M} \\
\hline
\rowcolor{gray!25}
 & 27.90 & 14.55 & 17.46 & 7.51 & 15.41 & 16.90 & 8.73 & \textbf{7.11} & 8.38 \\
\rowcolor{gray!25}
\multirow{-2}{*}{N} & \tiny(16.17) & \tiny(7.19) & \tiny(4.27) & \tiny(5.49) & \tiny(4.40) & \tiny(3.09) & \tiny(3.23) & \textbf{\tiny(3.58)} & \tiny(2.79) \\
 & 26.77 & 25.65 & 21.20 & 12.42 & 13.92 & 22.46 & 13.16 & \textbf{8.50} & 10.13 \\
\multirow{-2}{*}{I} & \tiny(3.61) & \tiny(5.00) & \tiny(2.12) & \tiny(4.85) & \tiny(4.72) & \tiny(3.78) & \tiny(3.72) & \textbf{\tiny(3.46)} & \tiny(2.62) \\
\rowcolor{gray!25}
 & 12.38 & 12.26 & 14.91 & 24.51 & 26.02 & 20.84 & 12.80 & \textbf{4.55} & 5.78 \\
\rowcolor{gray!25}
\multirow{-2}{*}{A} & \tiny(6.05) & \tiny(6.00) & \tiny(7.00) & \tiny(11.24) & \tiny(7.52) & \tiny(5.32) & \tiny(3.63) & \textbf{\tiny(3.19)} & \tiny(2.64) \\
 & 15.40 & 12.26 & 17.15 & 14.81 & 18.80 & 17.95 & 9.24 & \textbf{6.70} & 10.58 \\
\multirow{-2}{*}{B} & \tiny(4.24) & \tiny(3.81) & \tiny(2.46) & \tiny(7.85) & \tiny(5.22) & \tiny(3.06) & \tiny(3.08) & \textbf{\tiny(3.12)} & \tiny(2.78) \\
\rowcolor{gray!25}
 & 14.86 & 12.64 & 11.79 & 12.02 & 15.25 & 11.32 & 4.26 & \textbf{2.03} & 10.77 \\
\rowcolor{gray!25}
\multirow{-2}{*}{C} & \tiny(4.82) & \tiny(4.63) & \tiny(2.09) & \tiny(5.71) & \tiny(3.92) & \tiny(2.07) & \tiny(2.30) & \textbf{\tiny(1.63)} & \tiny(3.92) \\
 & 8.42 & 49.30 & 12.29 & \multirow{2}{*}{--} & \multirow{2}{*}{--} & 10.96 & 5.62 & \textbf{3.23} & 5.50 \\
\multirow{-2}{*}{H} & \tiny(3.18) & \tiny(1.75) & \tiny(2.93) &  &  & \tiny(2.29) & \tiny(2.47) & \textbf{\tiny(4.15)} & \tiny(6.68) \\
\rowcolor{gray!25}
 & 11.75 & 14.05 & 15.05 & 15.84 & 17.40 & 13.02 & 8.09 & 5.32 & \textbf{3.13} \\
\rowcolor{gray!25}
\multirow{-2}{*}{L} & \tiny(7.40) & \tiny(6.27) & \tiny(5.50) & \tiny(7.98) & \tiny(5.86) & \tiny(4.04) & \tiny(3.82) & \tiny(3.31) & \textbf{\tiny(1.52)} \\
 & 50.82 & 10.82 & 11.05 & 12.26 & 19.33 & 13.90 & 7.05 & 2.92 & \textbf{2.29} \\
\multirow{-2}{*}{P} & \tiny(5.08) & \tiny(6.11) & \tiny(3.87) & \tiny(6.90) & \tiny(5.37) & \tiny(4.33) & \tiny(3.46) & \tiny(2.29) & \textbf{\tiny(1.53)} \\
\rowcolor{gray!25}
 & 24.19 & 14.46 & 15.44 & 6.46 & 10.47 & 7.86 & 4.17 & \textbf{2.64} & 11.64 \\
\rowcolor{gray!25}
\multirow{-2}{*}{R} & \tiny(11.21) & \tiny(4.73) & \tiny(2.94) & \tiny(4.77) & \tiny(3.07) & \tiny(2.13) & \tiny(2.57) & \textbf{\tiny(2.09)} & \tiny(3.65) \\
 & 12.12 & 21.98 & 16.20 & 5.45 & 8.77 & 8.56 & 3.23 & \textbf{2.23} & 5.01 \\
\multirow{-2}{*}{W} & \tiny(5.31) & \tiny(4.86) & \tiny(2.37) & \tiny(3.91) & \tiny(4.28) & \tiny(1.97) & \tiny(2.46) & \textbf{\tiny(1.87)} & \tiny(2.44) \\
\hline
$\overline{\textrm{rank}}$ & 6.9 & 6.0 & 6.6 & 5.3 & 7.5 & 5.9 & 3.0 & 1.2 & 2.6 \\
\hline
\end{tabular}
\caption{Mean absolute error (standard deviation in parentheses), in percentages, for experiment \#2.}
\label{tab:err:xp:rnd}
\end{table}

\end{appendices}